# Efficient Methods for Qualitative Spatial Reasoning

**Jochen Renz** RENZ@DBAI.TUWIEN.AC.AT
*Institut für Informationssysteme, Technische Universität Wien*
*Favoritenstr.9, A-1040 Wien, Austria*

**Bernhard Nebel** NEBEL@INFORMATIK.UNI-FREIBURG.DE
*Institut für Informatik, Albert-Ludwigs-Universität*
*Am Flughafen 17, D-79110 Freiburg, Germany*

## Abstract

The theoretical properties of qualitative spatial reasoning in the RCC-8 framework have been analyzed extensively. However, no empirical investigation has been made yet. Our experiments show that the adaption of the algorithms used for qualitative temporal reasoning can solve large RCC-8 instances, even if they are in the *phase transition* region – provided that one uses the maximal tractable subsets of RCC-8 that have been identified by us. In particular, we demonstrate that the orthogonal combination of heuristic methods is successful in solving almost all apparently hard instances in the phase transition region up to a certain size in reasonable time.

## 1. Introduction

Representing qualitative spatial information and reasoning with such information is an important subproblem in many applications, such as natural language understanding, document interpretation, and geographical information systems. The RCC-8 calculus (Randell, Cui, & Cohn, 1992b) is well suited for representing topological relationships between spatial regions. Inference in the full calculus is, however, NP-hard (Grigni, Papadias, & Papadimitriou, 1995; Renz & Nebel, 1999). While this means that it is unlikely that very large instances can be solved in reasonable time, this result does not rule out the possibility that we can solve instances up to a certain size in reasonable time. Recently, maximal tractable subsets of RCC-8 were identified (Renz & Nebel, 1999; Renz, 1999) which can be used to speed up backtracking search for the general NP-complete reasoning problem by reducing the search space considerably.

In this paper we address several questions that emerge from previous theoretical results on RCC-8 (Renz & Nebel, 1999; Renz, 1999): Up to which size is it possible to solve instances in reasonable time? Which heuristic is the best? Is it really so much more efficient to use the maximal tractable subsets for solving instances of the NP-complete consistency problem as the theoretical savings given by the smaller branching factors indicate or is this effect out-balanced by the forward-checking power of the interleaved path-consistency computations? This was the case for similar temporal problems (pointisable vs. ORD-Horn relations) (Nebel, 1997). Is it possible to combine the different heuristics in such a way that more instances can be solved in reasonable time than by each heuristic alone?

We treat these questions by randomly generating instances and solving them using different heuristics. In doing so, we are particularly interested in the hardest randomly generated instances which leads to the question of phase-transitions (Cheeseman, Kanefsky, & Taylor, 1991): Is there a parameter for randomly generating instances of the consistency problem of RCC-8 that results in a phase-transition behavior? If so, is it the case that the hardest instances are mainly located in the phase-transition region while the instances not contained in the phase-transition region are easily solvable? In order to generate instances which are harder with a higher probability, we generate two different kinds of instances. On the one hand we generated instances which contain constraints over all RCC-8 relations, on the other hand we generated instances which contain only constraints over relations which are not contained in any of the maximal tractable subsets. We expect these instances to be harder on average than the former instances.

The algorithmic techniques we use for solving these randomly generated instances are borrowed from similar work on qualitative temporal reasoning (Nebel, 1997; van Beek & Manchak, 1996; Ladkin & Reinefeld, 1992). Additionally, we make use of the fragments of RCC-8, named $\widehat{\mathcal{H}}_8$, $\mathcal{Q}_8$, and $\mathcal{C}_8$, that permit polynomial-time inferences (Renz & Nebel, 1999; Renz, 1999). In the backtracking algorithm, which is used to solve the reasoning problem for full RCC-8, we decompose every disjunctive relation into relations of one of these tractable subsets instead of decomposing them into its base relations. This reduces the average branching factor of the backtracking tree from 4.0 for the base relations to 1.4375 for $\widehat{\mathcal{H}}_8$, to 1.523 for $\mathcal{C}_8$, and to 1.516 for $\mathcal{Q}_8$. Although these theoretical savings cannot be observed in our experiments, using the maximal tractable subsets instead of the base relations leads to significant performance improvements.

This paper is structured as follows. In Section 2, we give a brief sketch of the RCC-8 calculus and of the algorithms used for solving instances of RCC-8. In Section 3 we describe the procedure for randomly generating instances, the different heuristics we apply for solving these instances, and how we measure the quality of the heuristics. In Section 4 we evaluate different path-consistency algorithms in order to find the most efficient one to be used for forward-checking in the backtracking search. In Section 5 we observe a phase-transition behavior of the randomly generated instances and show that the instances in the phase-transition region are harder to solve than the other instances. In Section 6 we report on the outcome of running the different heuristics for solving the instances and identify several hard instances which are mainly located in the phase-transition region. In Section 7 we try to solve the hard instances by orthogonally combining the different heuristics. This turns out to be very effective and leads to a very efficient solution strategy. Finally, in Section 8 we evaluate this strategy by trying to solve very large instances.[1]

## 2. The Region Connection Calculus RCC-8

The Region Connection Calculus (RCC) is a first-order language for representation of and reasoning about topological relationships between extended spatial regions (Randell et al., 1992b). Spatial regions in RCC are non-empty regular subsets of some topological space which do not have to be internally connected, i.e., a spatial region may consist of different disconnected pieces. Different relationships between spatial regions can be defined based on one dyadic relation, the *connected* relation $\mathsf{C}(a,b)$ which is true if the topological closures of the spatial regions $a$ and $b$ share a common point.

1. The programs are available as an online appendix.

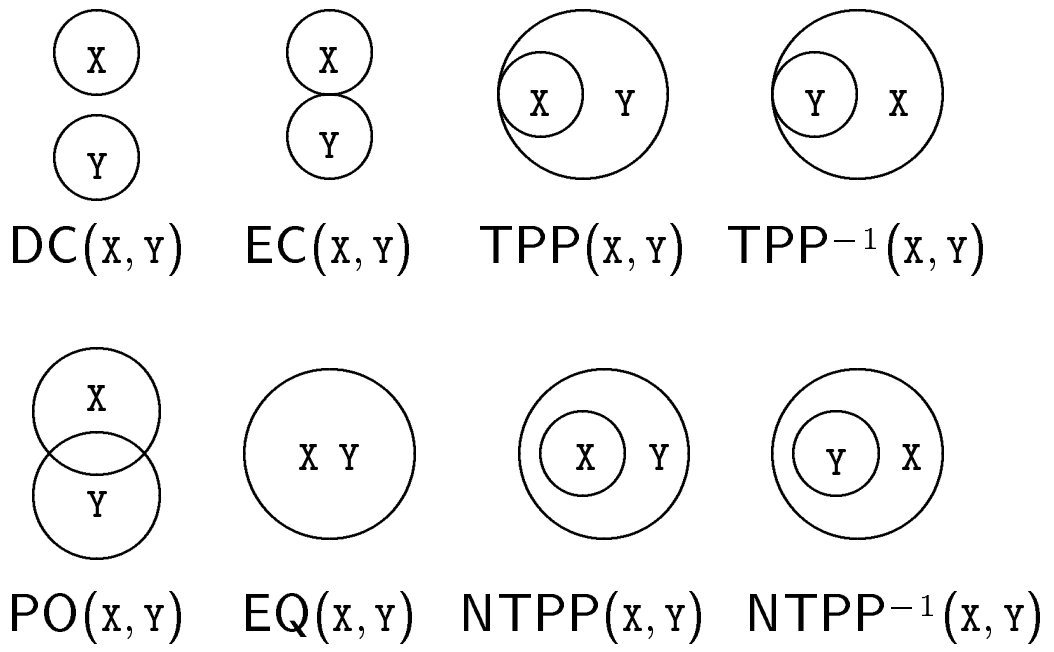

Figure 1: Two-dimensional examples for the eight base relations of RCC-8

The Region Connection Calculus RCC-8 is a constraint language formed by the eight jointly exhaustive and pairwise disjoint *base relations* DC, EC, PO, EQ, TPP, NTPP, $\mathsf{TPP}^{-1}$, and $\mathsf{NTPP}^{-1}$ definable in the RCC-theory and by all possible unions of the base relations—giving a total number of $2^8 = 256$ different relations. The base relations have the meaning of *DisConnected, Externally Connected, Partial Overlap, EQual, Tangential Proper Part, Non-Tangential Proper Part*, and their converses. Examples for these relations are shown in Figure 1. Constraints are written in the form $xRy$ where $x, y$ are variables for spatial regions and $R$ is an RCC-8 relation. We write the union of base relations as $\{R, S\}$. The union of all base relations, the *universal relation*, is written as $\{*\}$. Apart from union ($\cup$), other operations on relations are defined, namely, converse ($^{\smile}$), intersection ($\cap$), and composition ($\circ$). The formal definitions of these operations are:

$$\begin{array}{llcl}\forall x, y: & x(R \cup S)y & \leftrightarrow & xRy \vee xSy, \\ \forall x, y: & x(R \cap S)y & \leftrightarrow & xRy \wedge xSy, \\ \forall x, y: & xR^{\smile}y & \leftrightarrow & yRx, \\ \forall x, y: & x(R \circ S)y & \leftrightarrow & \exists z : (xRz \wedge zSy).\end{array}$$

The composition of base relations can be computed from the semantics of the relations and is usually provided as a composition table (Randell, Cohn, & Cui, 1992a; Bennett, 1994). The RCC-8 composition table corresponds to the given extensional definition of composition only if the universal region is not permitted (Bennett, 1997). Based on this table, compositions of disjunctive relations can be easily computed. In the following, $\widehat{\mathcal{S}}$ denotes the closure of a set of RCC-8 relations $\mathcal{S}$ under composition, intersection, and converse.

A finite set of RCC-8 constraints $\Theta$ describing the topological relationships of $n$ different regions can be represented by an $n \times n$ matrix $M$, where each entry $M_{ij}$ represents the RCC-8 relation holding between region $i$ and region $j$. Without loss of generality, $M_{ii} = \{\mathsf{EQ}\}$ and $M_{ji} = M_{ij}^{\smile}$ can be assumed. The fundamental reasoning problem (named RSAT) in this framework is deciding *consistency* of a set of spatial formulas $\Theta$, i.e., whether there is a spatial configuration where the relations between the regions can be described by $\Theta$. All other interesting reasoning problem can be reduced to it in polynomial time (Golumbic & Shamir, 1993). Unfortunately, RSAT is NP-complete (Renz & Nebel, 1999), i.e., it is unlikely that there is any polynomial algorithm for deciding consistency. However, it was shown in Nebel's (1995) paper that there are subsets $\mathcal{S}$ of RCC-8 for which the consistency problem (written $\mathsf{RSAT}(\mathcal{S})$) can be decided in polynomial time.[2] In particular the set of eight base relations $\mathcal{B}$ was shown to be tractable. From that it follows that $\widehat{\mathcal{B}}$ consisting of 32 relations is also tractable. An even larger tractable subset containing all base relations is $\widehat{\mathcal{H}}_8$ (Renz & Nebel, 1999), which contains 148 out of the 256 RCC-8 relations. This set was also shown to be maximal with respect to tractability, i.e., if any other RCC-8 relation is added, the consistency problem becomes NP-complete. Renz (1999) made a complete analysis of tractability of RSAT by identifying all maximal tractable subsets which contain all base relations, altogether three subsets $\widehat{\mathcal{H}}_8$, $\mathcal{Q}_8$ (160 relations), and $\mathcal{C}_8$ (158 relations). $\mathcal{NP}_8$ is the set of relations that by themselves result in NP-completeness when combined with the set of base relations. It contains the following 76 relations which are not contained in one of $\widehat{\mathcal{H}}_8, \mathcal{Q}_8$, or $\mathcal{C}_8$ (Renz, 1999):

$$\begin{aligned} \mathcal{NP}_8 &= \{R \mid \{\mathsf{PO}\} \not\subseteq R \text{ and } (\{\mathsf{NTPP}\} \subseteq R \text{ or } \{\mathsf{TPP}\} \subseteq R) \\ &\qquad \text{and } (\{\mathsf{NTPP}^{-1}\} \subseteq R \text{ or } \{\mathsf{TPP}^{-1}\} \subseteq R)\} \\ &\quad \cup \{\{\mathsf{EC}, \mathsf{NTPP}, \mathsf{EQ}\}, \{\mathsf{DC}, \mathsf{EC}, \mathsf{NTPP}, \mathsf{EQ}\}, \\ &\qquad \{\mathsf{EC}, \mathsf{NTPP}^{-1}, \mathsf{EQ}\}, \{\mathsf{DC}, \mathsf{EC}, \mathsf{NTPP}^{-1}, \mathsf{EQ}\}\}. \end{aligned}$$

The maximal tractable subsets contain the following relations (Renz, 1999):

$$\begin{aligned} \widehat{\mathcal{H}}_8 &= (\mathsf{RCC\text{-}8} \setminus \mathcal{NP}_8) \setminus \{R \mid (\{\mathsf{EQ}, \mathsf{NTPP}\} \subseteq R \text{ and } \{\mathsf{TPP}\} \not\subseteq R) \\ &\qquad \text{or } (\{\mathsf{EQ}, \mathsf{NTPP}^{-1}\} \subseteq R \text{ and } \{\mathsf{TPP}^{-1}\} \not\subseteq R)\} \\ \mathcal{C}_8 &= (\mathsf{RCC\text{-}8} \setminus \mathcal{NP}_8) \setminus \{R \mid \{\mathsf{EC}\} \subset R \text{ and } \{\mathsf{PO}\} \not\subseteq R \text{ and} \\ &\qquad R \cap \{\mathsf{TPP}, \mathsf{NTPP}, \mathsf{TPP}^{-1}, \mathsf{NTPP}^{-1}, \mathsf{EQ}\} \neq \emptyset\} \\ \mathcal{Q}_8 &= (\mathsf{RCC\text{-}8} \setminus \mathcal{NP}_8) \setminus \{R \mid \{\mathsf{EQ}\} \subset R \text{ and } \{\mathsf{PO}\} \not\subseteq R \text{ and} \\ &\qquad R \cap \{\mathsf{TPP}, \mathsf{NTPP}, \mathsf{TPP}^{-1}, \mathsf{NTPP}^{-1}\} \neq \emptyset\} \end{aligned}$$

All relations of $\mathcal{Q}_8$ are contained in one of $\widehat{\mathcal{H}}_8$ or $\mathcal{C}_8$, i.e., $\widehat{\mathcal{H}}_8 \cup \mathcal{C}_8 = \mathsf{RCC\text{-}8} \setminus \mathcal{NP}_8$. Although $\widehat{\mathcal{H}}_8$ is the smallest of the three maximal tractable subsets, it best decomposes the RCC-8 relations: When decomposing an RCC-8 relation $R$ into sub-relations $S_i$ of one of the maximal tractable subsets, i.e., $R = S_1 \cup \ldots \cup S_k$, one needs on average 1.4375 $\widehat{\mathcal{H}}_8$ relations, 1.516 $\mathcal{Q}_8$ relations, and 1.523 $\mathcal{C}_8$ relations for decomposing all RCC-8 relations. Renz (2000) gives a detailed enumeration of the relations of the three sets.

### 2.1 The Path-Consistency Algorithm

As in the area of qualitative temporal reasoning based on Allen's interval calculus (Allen, 1983), the *path-consistency algorithm* (Montanari, 1974; Mackworth, 1977; Mackworth & Freuder, 1985) can be used to approximate consistency and to realize *forward-checking* (Haralick & Elliot, 1980) in a backtracking algorithm.

The path-consistency algorithm checks the consistency of all triples of relations and eliminates relations that are impossible. This is done by iteratively performing the following operation

$$M_{ij} \leftarrow M_{ij} \cap M_{ik} \circ M_{kj}$$

2. Strictly speaking, this applies only to systems of regions that do not require regularity.

*Algorithm*: Path-consistency
*Input*: A set $\Theta$ of binary constraints over the variables $x_1, x_2, \ldots, x_n$ of $\Theta$ represented by an $n \times n$ matrix $M$.
*Output*: path-consistent set equivalent to $\Theta$; `fail`, if such a set does not exist.

1. $Q := \{(i,j,k),(k,i,j) \mid 1 \leq i,j,k \leq n, i < j, k \neq i, k \neq j\}$;
   ($i$ indicates the $i$-th variable of $\Theta$. Analogously for $j$ and $k$)
2. *while* $Q \neq \emptyset$ *do*
3. select and delete a path $(p,r,q)$ from $Q$;
4. *if* REVISE$(p,r,q)$ *then*
5. *if* $M_{pq} = \emptyset$ *then return* `fail`
6. *else* $Q := Q \cup \{(p,q,s),(s,p,q) \mid 1 \leq s \leq n, s \neq p, s \neq q\}$.

*Function*: REVISE$(i,k,j)$
*Input*: three labels $i$, $k$ and $j$ indicating the variables $x_i, x_j, x_k$ of $\Theta$
*Output*: `true`, if $M_{ij}$ is revised; `false` otherwise.
*Side effects*: $M_{ij}$ and $M_{ji}$ revised using the operations $\cap$ and $\circ$ over the constraints involving $x_i$, $x_k$, and $x_j$.

1. oldM := $M_{ij}$;
2. $M_{ij} := M_{ij} \cap (M_{ik} \circ M_{kj})$;
3. *if* (oldM $= M_{ij}$) *then return* `false`;
4. $M_{ji} := M_{ij}^{\smile}$;
5. *return* `true`.

Figure 2: Path-consistency algorithm.

for all triples of regions $i,j,k$ until a fixed point $\overline{M}$ is reached. If $\overline{M}_{ij} = \emptyset$ for a pair $i,j$, then we know that $M$ is inconsistent, otherwise $\overline{M}$ is *path-consistent*. Computing $\overline{M}$ can be done in $O(n^3)$ time (see Figure 2). This is achieved by using a *queue* of triples of regions for which the relations should be recomputed (Mackworth & Freuder, 1985). Path-consistency does not imply consistency. For instance, the following set of spatial constraints is path-consistent but not consistent:

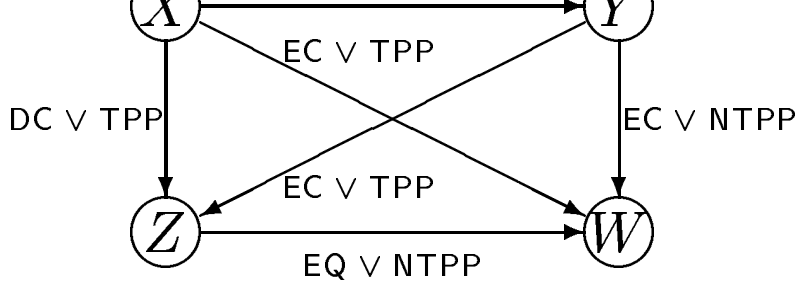

On the other hand, consistency does not imply path-consistency, since path-consistency is not a form of consistency (in its logical sense), but a form of disjunctive non-redundancy. Nevertheless, path-consistency can be enforced to any consistent set of constraints by ap-

*Algorithm*: Consistency
*Input*: A set $\Theta$ of RCC-8 constraints over the variables $x_1, x_2, \ldots, x_n$ and a subset $\mathcal{S} \subseteq$ RCC-8 that contains all base relations and for which Decide is a sound and complete decision procedure.
*Output*: `true`, iff $\Theta$ is consistent.

1. Path-Consistency($\Theta$)
2. *if* $\Theta$ contains the empty relation *then return* `false`
3. *else* choose an unprocessed constraint $x_i R x_j$ and split $R$ into $S_1, \ldots, S_k \in \mathcal{S}$ such that $S_1 \cup \ldots \cup S_k = R$
4. *if* no constraint can be split *then return* Decide($\Theta$)
5. *for* all refinements $S_l$ $(1 \leq l \leq k)$ *do*
6. replace $x_i R x_j$ with $x_i S_l x_j$ in $\Theta$
7. *if* Consistency($\Theta$) *then return* `true`

Figure 3: Backtracking algorithm for deciding consistency.

plying a path-consistency algorithm. If only relations in $\widehat{\mathcal{H}}_8$, $\mathcal{Q}_8$, or $\mathcal{C}_8$ are used, however, the path-consistency algorithm is sufficient for deciding consistency, i.e., path-consistency decides RSAT($\widehat{\mathcal{H}}_8$), RSAT($\mathcal{Q}_8$), and RSAT($\mathcal{C}_8$), (Renz & Nebel, 1999; Renz, 1999).

### 2.2 The Backtracking Algorithm

In order to solve an instance $\Theta$ of RSAT, we have to explore the corresponding search space using some sort of backtracking. In our experiments, we used a backtracking algorithm employed for solving qualitative temporal reasoning problems (Nebel, 1997), which is based on the algorithm proposed by Ladkin and Reinefeld (1992). For this algorithm (see Figure 3) it is necessary to have a subset $\mathcal{S} \subseteq$ RCC-8 for which consistency can be decided by using a sound and complete (and preferably polynomial) decision procedure Decide. If $\mathcal{S}$ contains all base relations, then each relation $R \in$ RCC-8 can be decomposed into sub-relations $S_i \in \mathcal{S}$ such that $R = \bigcup_i S_i$. The size of a particular decomposition is the minimal number of sub-relations $S_i$ which is used to decompose $R$. The backtracking algorithm successively selects constraints of $\Theta$, backtracks over all sub-relations of the constraints according to their decomposition and decides sub-instances which contain only constraints over $\mathcal{S}$ using Decide.

The (optional) procedure Path-consistency in line 1 is used for forward-checking and restricts the remaining search space. Nebel (1997) showed that this restriction does not effect soundness and completeness of the algorithm. If enforcing path-consistency is sufficient for deciding RSAT($\mathcal{S}$), Decide($\Theta$) in line 5 is not necessary. Instead it is possible to always return `true` there.

The efficiency of the backtracking algorithm depends on several factors. One of them is, of course, the size of the search space which has to be explored. A common way of measuring the size of the search space is the average *branching factor* $b$ of the search space, i.e., the average number of branches each node in the search space has (a node is a recursive call of Consistency). Then the average size of the search space can be computed as $b^{(n^2-n)/2}$, where $(n^2-n)/2$ is the number of constraints which have to be split when $n$ variables are given. For the backtracking algorithm described in Figure 3 the branching factor depends on the average number of relations of the split set $\mathcal{S}$ into which a relation has to be split. The less splits on average the better, i.e., it is to be expected that the efficiency of the backtracking algorithm depends on the split set $\mathcal{S}$ and its branching factor. Another factor is how the search space is explored. The backtracking algorithm of Figure 3 offers two possibilities for applying heuristics. One is in line 3 where the next unprocessed constraint can be chosen, the other is in line 5 where the next refinement can be chosen. These two choices influence the search space and the path through the search space.

## 3. Test Instances, Heuristics, and Measurement

There is no previous work on empirical evaluation of algorithms for reasoning with RCC-8 and no benchmark problems are known. Therefore we randomly generated our test instances with a given number of regions $n$, an average label-size $l$, and an average degree $d$ of the constraint graph. Further, we used two different sets of relations for generating test instances, the set of all RCC-8 relations and the set of hard RCC-8 relations $\mathcal{NP}_8$, i.e., those 76 relations which are not contained in any of the maximal tractable subsets $\widehat{\mathcal{H}}_8$, $\mathcal{C}_8$, or $\mathcal{Q}_8$. Based on these sets of relations, we used two models to generate instances, denoted by $A(n,d,l)$ and $H(n,d,l)$. The former model uses all relations to generate instances, the latter only the relations in $\mathcal{NP}_8$. The instances are generated as follows:

1. A constraint graph with $n$ nodes and an average degree of $d$ for each node is generated. This is accomplished by selecting $nd/2$ out of the $n(n-1)/2$ possible edges using a uniform distribution.

2. If there is no edge between the $i$th and $j$th node, we set $M_{ij} = M_{ji}$ to be the universal relation.

3. Otherwise a non-universal relation is selected according to the parameter $l$ such that the average size of relations for selected edges is $l$. This is accomplished by selecting one of the base relations with uniform distribution and out of the remaining 7 relations each one with probability $(l-1)/7$.[3] If this results in an allowed relation (i.e., a relation of $\mathcal{NP}_8$ for $H(n,d,l)$, any RCC-8 relation for $A(n,d,l)$), we assign this relation to the edge. Otherwise we repeat the process.

The reason for also generating instances using only relations of $\mathcal{NP}_8$ is that we assume that these instances are difficult to solve since every relation has to be split during the backtracking search, even if we use a maximal tractable subclass as the split set. We only generated instances of average label size $l = 4.0$, since in this case the relations are equally distributed.

3. This method could result in the assignment of a universal constraint to a selected link, thereby changing the degree of the node. However, since the probability of getting the universal relation is very low, we ignore this in the following.

This way of generating random instances is very similar to the way random CSP instances over finite domains are usually generated (Gent, MacIntyre, Prosser, Smith, & Walsh, 2001). Achlioptas et al. (1997) found that the standard models for generating random CSP instances over finite domains lead to trivially flawed instances for $n \to \infty$, i.e., instances become locally inconsistent without having to propagate constraints. Since we are using CSP instances over infinite domains, Achlioptas et al.'s result does not necessarily hold for our random instances. We, therefore, analyze in the following whether our instances are also trivially flawed for $n \to \infty$. In order to obtain a CSP over a finite domain, we first have to transform our constraint graph into its dual graph where each of the $n(n-1)/2$ edges $M_{ij}$ of our constraint graph corresponds to a node in the dual graph. Moreover, each of the $n$ variables of the constraint graph corresponds to $n-1$ edges in the dual graph, i.e., the dual graph contains $n(n-1)$ edges and $n(n-1)/2$ nodes. In the dual graph, each node corresponds to a variable over the eight-valued domain $\mathcal{D} = \{\mathsf{DC}, \mathsf{EC}, \mathsf{PO}, \mathsf{TPP}, \mathsf{TPP}^{-1}, \mathsf{NTPP}, \mathsf{NTPP}^{-1}, \mathsf{EQ}\}$. Ternary constraints over these variables are imposed by the composition table, i.e., the composition rules $M_{ij} \subseteq M_{ik} \circ M_{kj}$ must hold for all connected triples of nodes $M_{ij}, M_{ik}, M_{kj}$ of the dual graph ($M_{ij} = M_{ji}^{\smile}$ for all $i,j$). There are $\binom{n}{3} = n(n-1)(n-2)/6$ connected triples in the dual graph. The overall number of triples in the dual graph is $\binom{n(n-1)/2}{3}$. $nd/2$ unary constraints on the domain of the variables $M_{ij}$ are given, i.e., there are $\binom{nd/2}{3}$ triples in the dual graph where all nodes are restricted by unary constraints. Therefore, the expected number $E_{CT}^{n}$ of connected triples for which unary constraints are given can be computed as

$$E_{CT}^{n} = \frac{\binom{n}{3} \cdot \binom{nd/2}{3}}{\binom{n(n-1)/2}{3}}.$$

For $n \to \infty$, the expected number of triples $E_{CT}^{\infty}$ tends to $d^3/6$. For the instances generated according to the model $A(n,d,l)$, the probability that the unary constraints which are assigned to a triple lead to a local inconsistency is about $0,0036\%$ (only 58,989 out of the $255^3 = 16,581,375$ possible assignments are inconsistent). Since one locally inconsistent triple makes the whole instance inconsistent, we are interested in the average degree $d$ for which the expected number $E_{IT}^{n}$ of locally inconsistent triples is equal to one. For the model $A(n,d,l)$ this occurs for a value of $d = 11.90$, and $E_{IT}^{\infty} = 0.5$ for $d = 9.44$. For $n = 100$, the expected number of locally inconsistent triples is one for $d = 13.98$, and $E_{IT}^{100} = 0.5$ for $d = 11.10$. For the model $H(n,d,l)$, none of the possible assignments of the triples leads to a local inconsistency, i.e., all triples of the randomly generated instances of the $H(n,d,l)$ model are locally consistent.[4] This analysis shows that contrary to what Achlioptas et al. found for randomly generated CSP instances over finite domains, the model $H(n,d,l)$, and the model $A(n,d,l)$ for $d$ small do not suffer from trivial local inconsistencies.

4. This is similar to the result for CSPs over finite domains that by restricting the constraint type, e.g., if only "not-equal" constraints as in graph-coloring are used, it is possible to ensure that problems cannot be trivially flawed.

We solve the randomly generated instances using the backtracking algorithm described in the previous section. The search space on which backtracking is performed depends on the split set, i.e., the set of sub-relations that is allowed in the decompositions. Choosing the right split-set influences the search noticeably as it influences the average branching factor of the search space. We choose five different split sets, the three maximal tractable subsets $\widehat{\mathcal{H}}_8$, $\mathcal{Q}_8$, and $\mathcal{C}_8$, the set of base relations $\mathcal{B}$ and the closure of this set $\widehat{\mathcal{B}}$ which consists of 38 relations. These sets have the following branching factors $\mathcal{B}$: 4.0, $\widehat{\mathcal{B}}$: 2.50 ,$\widehat{\mathcal{H}}_8$: 1.438, $\mathcal{C}_8$: 1.523, $\mathcal{Q}_8$: 1.516. This is, of course, a worst case measure because the interleaved path-consistency computations reduce the branching factor considerably (Ladkin & Reinefeld, 1997).

Apart from the choice of the split set there are other heuristics which influence the efficiency of the search. In general it is the best search strategy to proceed with the constraint with the most constraining relation (line 3 of Figure 3) and the least constraining choice of a sub-relation (line 5 of Figure 3). We investigated two different aspects for choosing the next constraint to be processed (Nebel, 1997).

**static/dynamic:** Constraints are processed according to a heuristic evaluation of their constrainedness which is determined *statically* before the backtracking starts or *dynamically* during the search.

**local/global:** The evaluation of the constrainedness is based on a *local* heuristic weight criterion or on a *global* heuristic criterion (van Beek & Manchak, 1996).

This gives us four possibilities we can combine with the five different split sets, i.e., a total number of 20 different heuristics. The evaluation of constrainedness as well as how relations are decomposed into relations of different split sets depends on the *restrictiveness* of relations, which is a heuristic criterion (van Beek & Manchak, 1996). Restrictiveness of a relation is a measure of how a relation restricts its neighborhood. For instance, the universal relation given in a constraint network does not restrict its neighboring relations at all, the result of the composition of any relation with the universal relation is the universal relation. The identity relation, in contrast, restricts its neighborhood a lot. In every triple of variables where one relation is the identity relation, the other two relations must be equal. Therefore, the universal relation is usually the least restricting relation, while the identity relation is usually the most restricting relation. Restrictiveness of relations is represented as a weight in the range of 1 to 16 assigned to every relation, where 1 is the value of the most and 16 the value of the least restricting relation. We discuss in the following section in detail how the restrictiveness and the weight of a relation is determined.

Given the weights assigned to every relation, we compute decompositions and estimate constrainedness as follows. For each split set $\mathcal{S}$ and for each RCC-8 relation $R$ we compute the smallest decomposition of $R$ into sub-relations of $\mathcal{S}$, i.e., the decomposition which requires the least number of sub-relations of $\mathcal{S}$. If there is more than one possibility, we choose the decomposition with the least restricting sub-relations. In line 5 of the backtracking algorithm (see Figure 3), the least restricting sub-relation of each decomposition is processed first. For the local strategy, the constrainedness of a constraint is determined by the size of its decomposition (which can be different for every split set) and by its weight. We choose the constraint with the smallest decomposition larger than one and, if there is more than one such constraint, the one with the smallest weight. The reason for choosing the relation with the smallest decomposition is that it is expected that forward-checking refines relations with a larger decomposition into relations with a smaller decomposition. This reduces the backtracking effort. For the global strategy, the constrainedness of a constraint $xRy$ is determined by adding the weights of all neighboring relations $S, T$ with $xSz$ and $zTy$ to the weight of $R$. The idea behind this strategy is that when refining the relation $R$ with the most restricted neighborhood, an inconsistency is detected faster than when refining a relation with a less restricted neighborhood.

In order to evaluate the quality of the different heuristics, we measured the run-time used for solving instances as well as the number of visited nodes in the search space. Comparing different approaches by their run-time is often not very reliable as it depends on several factors such as the implementation of the algorithms, the used hardware, or the current load of the used machine which makes results sometimes not reproducible. For this reason, we ran all our run-time experiments on the same machine, a Sun Ultra 1 with 128 MB of main memory. Nevertheless, we suggest to use the run-time results mainly for qualitatively comparing different heuristics and for getting a rough idea of the order of magnitude for which instances can be solved.

In contrast to this, the number of visited nodes for solving an instance with a particular heuristic is always the same on every machine. This allows comparing the path through the search space taken by the single heuristics and to judge which heuristic makes the better choices on average. However, this does not take into account the time that is needed to make a choice at a single node. Computing the local constrainedness of a constraint is certainly faster than computing its global constrainedness. Similarly, computing constrainedness statically should be faster than computing it dynamically. Furthermore, larger instances require more time at the nodes than smaller instances, be it for computing path-consistency or for computing the constrainedness. Taking running-time and the number of visited nodes together gives good indications of the quality of the heuristics.

A further choice we make in evaluating our measurements is that of how to aggregate the measurements of the single instances to a total picture. Some possibilities are to use either the average or different percentiles such as the median, i.e., the 50% percentile. The $d\%$ percentile for a value $0 < d < 100$ is obtained by sorting the measurements in increasing order and picking the measurement of the $d\%$ element, i.e., $d\%$ of the values are less than that value. Suppose that most instances have a low value (e.g. running time) and only a few instances have a very large value. Then the average might be larger than the values of almost all instances, while in this case the median is a better indication of the distribution of the values. In this case the 99% percentile, for instance, gives a good indication of the value of the hardest among the "normal" instances. We have chosen to use the average value when the measurements are well distributed and to use both 50% and 99% percentile when there are only a few exceptional values in the distribution of the measurements.

## 4. Empirical Evaluation of the Path-Consistency Algorithm

Since the efficiency of the backtracking algorithm depends on the efficiency of the underlying path-consistency algorithm, we will first compare different implementations of the path-consistency algorithm. In previous empirical investigations (van Beek & Manchak, 1996) of reasoning with Allen's interval relations (Allen, 1983), different methods for computing the composition of two relations were evaluated. This was mainly because the full composition table for the interval relations contains $2^{13} \times 2^{13} = 67108864$ entries, which was too large at that time to be stored in the main memory. In our setting, we simply use a composition table that specifies the compositions of all RCC-8 relations, which is a $256 \times 256$ table consuming approximately 128 KB of main memory. This means that the composition of two arbitrary relations is done by a simple table lookup.

Van Beek and Manchak (1996) also studied the effect of weighting the relations in the queue according to their restrictiveness and process the most restricting relation first. Restrictiveness was measured for each base relation by successively composing the base relation with every possible label, summing up the cardinalities, i.e., the number of base relations contained in the result of the composition, and suitably scaling the result. The reason for doing so is that the most restricting relation restricts the other relations on average most and therefore decreases the probability that they have to be processed again. Restrictiveness of a complex relation was approximated by summing up the restrictiveness of the involved base relations. Van Beek and Manchak (1996) found that their method of weighting the triples in the queue is much more efficient than randomly picking an arbitrary triple. Because of the relatively small number of RCC-8 relations, we computed the exact restrictiveness by composing each relation with every other relation and summing up the cardinalities of the resulting compositions. We scaled the result into weights from 1 (the most restricting relation) to 16 (the least restricting relations).

This gives us three different implementations of the path-consistency algorithm. One in which the entries in the queue are not weighted, one with approximated restrictiveness as done by van Beek and Manchak, and one with exact restrictiveness.[5] In order to compare these implementations, we randomly generated instances with 50 to 1,000 regions. For each value of the average degree ranging from 8.0 stepping with 0.5 to 11.0 we generated 10 different instances. Figure 4 displays the average CPU time of the different methods for applying the path-consistency algorithm to the same generated instances. It can be seen that the positive effect of using a weighted queue is much greater for our problem than for the temporal problem (about 10× faster than using an ordinary queue without weights compared to only about 2× faster (van Beek & Manchak, 1996)). Determining the weights of every relation using their exact restrictiveness does not have much advantage over approximating their restrictiveness using the approach by van Beek and Manchak (1996), however. For our further experiments we always used the "exact weights" method because determining the restrictiveness amounts to just one table lookup.

As mentioned in the previous section, one way of measuring the quality of the heuristics is to count the number of visited nodes in the backtrack search. In our backtracking algorithm, path-consistency is enforced in every visited node. Note that it is not adequate to multiply the average running-time for enforcing path-consistency of an instance of a particular size with the number of visited nodes in order to obtain an approximation of the required running time for that instance. The average running-time for enforcing path-consistency as given in Figure 4 holds only when all possible paths are entered into the queue at the beginning of the computation (see line 1 of Figure 2). These are the paths

5. For the weighted versions we select a path $(i, k, j)$ from the queue $Q$ in line 3 of the algorithm of Figure 2 according to the weights of the different paths in $Q$ which are computed as specified above.

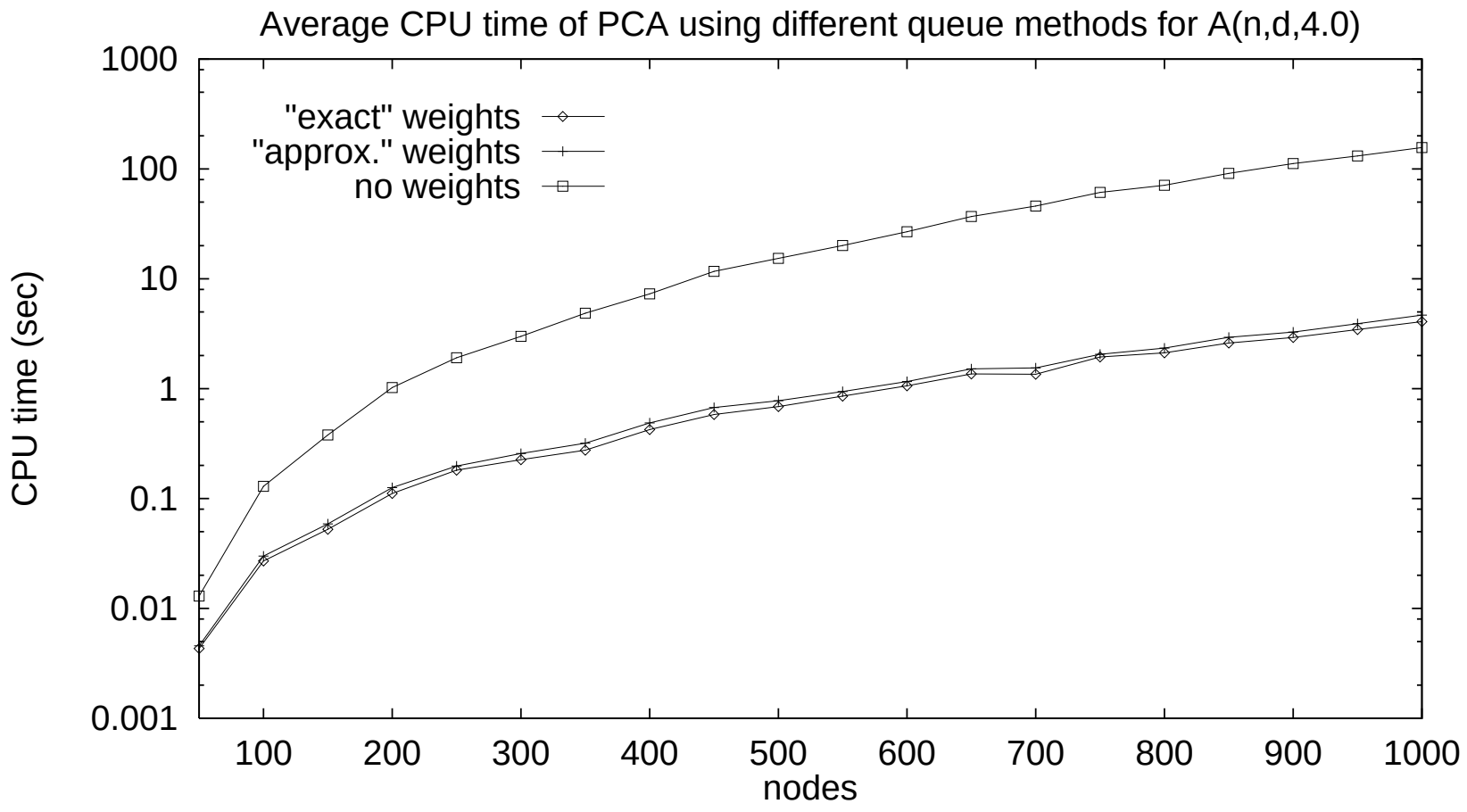

Figure 4: Comparing the performance of the path-consistency algorithm using different methods for weighting the queue (70 instances/data point, $d = 8.0 - 11.0$)

which have to be checked by the algorithm. The path-consistency computation during the backtracking search is different, however. There, only the paths involving the currently changed constraint are entered in the queue, since only these paths might result in changes of the constraint graph. This is much faster than the full computation of path-consistency which is only done once at the beginning of the backtrack search.

## 5. The Phase-Transition of RCC-8

When randomly generating problem instances there is usually a problem-dependent parameter which determines the solubility of the instances. In one parameter range instances are underconstrained and are therefore soluble with a very high probability. In another range, problems are overconstrained and soluble with a very low probability. In between these ranges is the *phase-transition region* where the probability of solubility changes abruptly from very high to very low values (Cheeseman et al., 1991). In order to study the quality of different heuristics and algorithms with randomly generated instances of an NP-complete problem, it is very important to be aware of the phase-transition behavior of the problem. This is because instances which are not contained in the phase-transition region are often very easily solvable by most algorithms and heuristics and are, thus, not very useful for comparing their quality. Conversely, hard instances which are better suited for comparing the quality of algorithms and heuristics are usually found in the phase-transition region.

In this section we identify the phase-transition region of randomly generated instances of the RSAT problem, both for instances using all RCC-8 relations and for instances using only relations of $\mathcal{NP}_8$. Similarly to the empirical analysis of qualitative temporal reasoning problems (Nebel, 1997), it turns out that the phase-transition depends most strongly on the average degree $d$ of the nodes in the constraint graph. If all relations are allowed, the phase-transition is around $d = 8$ to $d = 10$ depending on the instance size (see Figure 5). Because of the result of our theoretical analysis of the occurrence of trivial flaws (see Section 3), it can be expected that for larger instance sizes the phase-transition behavior will be overlaid and mainly determined by the expected number of locally inconsistent triples which also depends on the average degree $d$. Thus, although it seems that the phase-transition shifts towards larger values of $d$ as the instance size increases, the phase-transition is asymptotically below $d = 9.44$, the theoretical value for $n \to \infty$ (see Section 3). Instances which are not path-consistent can be solved very fast by just one application of the path-consistency algorithm without further need for backtracking. When looking at the median CPU times given in Figure 5, one notices that there is a sharp decline of the median CPU times at the phase transition. This indicates that for values of the average degree which are higher than where the phase-transition occurs, at least 50% of the instances are not path-consistent.

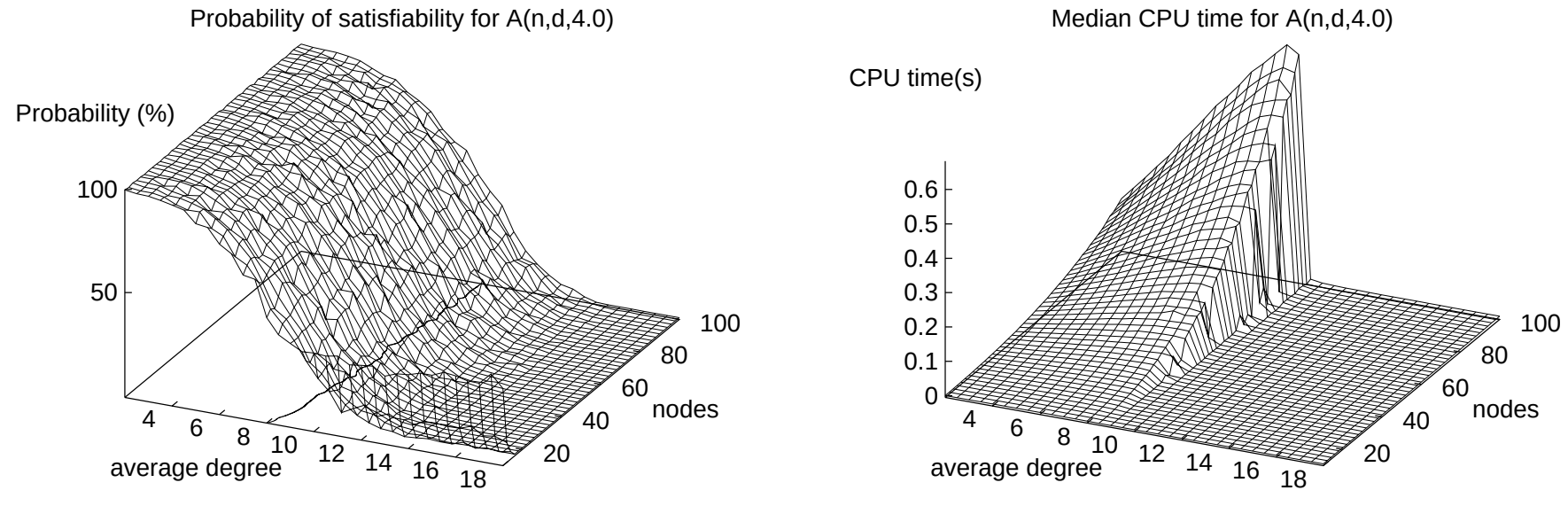

Figure 5: Probability of satisfiability and median CPU time for $A(n, d, 4.0)$ using the $\widehat{\mathcal{H}}_8$/static/global heuristic (500 instances per data point)

When using only "hard" relations, i.e., relations in $\mathcal{NP}_8$, the phase-transition appears at higher values for $d$, namely, between $d = 10$ and $d = 15$ (see Figure 6). As the median runtime shows, these instances are much harder in the phase-transition than in the former case. As in the previous case, but even more strongly, it seems that the phase-transition shifts towards larger values of $d$ as the instance size increases, and also that the phase-transition region narrows.

In order to evaluate the quality of the path-consistency method as an approximation to consistency, we counted the number of instances that are inconsistent but path-consistent (see Figure 7), i.e., those instances where the approximation of the path-consistency algorithm to consistency is wrong. First of all, one notes that all such instances are close to the phase transition region. In the general case, i.e., when constraints over all RCC-8 relations are employed, only a very low percentage of instances are path-consistent but inconsistent. Therefore, the figure looks very erratic. More data points would be required in order to obtain a smooth curve. However, a few important observations can be made from this figure, namely, that path-consistency gives an excellent approximation to consistency even for instances of a large size. Except for very few instances in the phase-transition region, almost all instances which are path-consistent are also consistent. This picture changes when looking at the $H(n, d, 4.0)$ case. Here almost all instances in the phase-transition region and many instances in the mostly insoluble region are path-consistent, though only a few of them are consistent.

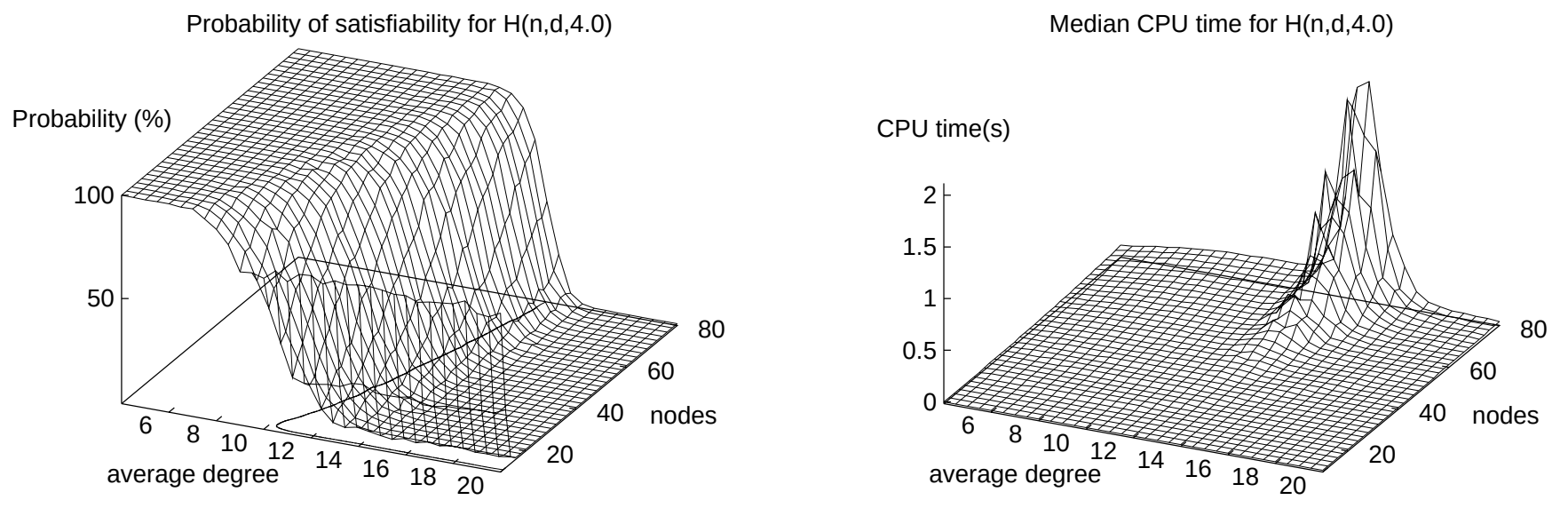

Figure 6: Probability of satisfiability and median CPU time for $H(n, d, 4.0)$ using the $\widehat{\mathcal{H}}_8$/static/global heuristic (500 instances per data point)

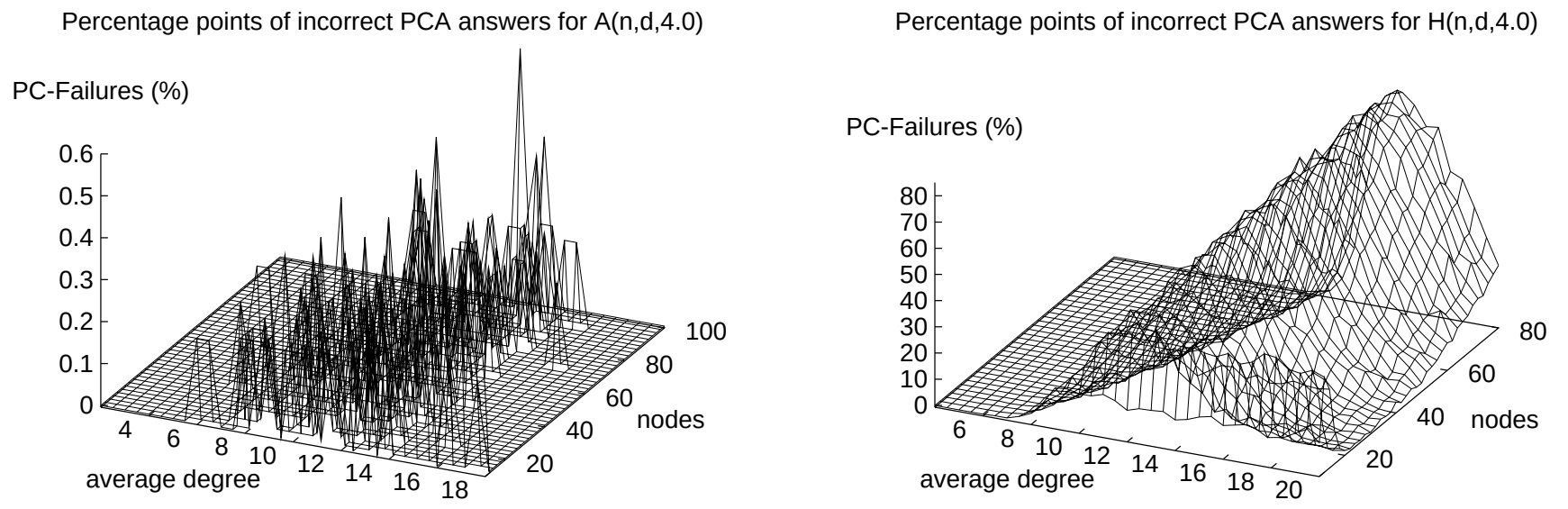

Figure 7: Percentage points of incorrect answers of the path-consistency algorithm for $A(n, d, 4.0)$ and $H(n, d, 4.0)$

For the following evaluation of the different heuristics we will randomly generate instances with an average degree between $d = 2$ and $d = 18$ in the $A(n, d, 4.0)$ case and between $d = 4$ and $d = 20$ in the $H(n, d, 4.0)$ case. This covers a large area around the phase-transition. We expect the instances in the phase-transition region of $H(n, d, 4.0)$ to be particularly hard which makes them very interesting for comparing the quality of the different heuristics.

## 6. Empirical Evaluation of the Heuristics

In this section we compare the different heuristics by running them on the same randomly generated instances. For the instances of $A(n, d, 4.0)$ we ran all 20 different heuristics

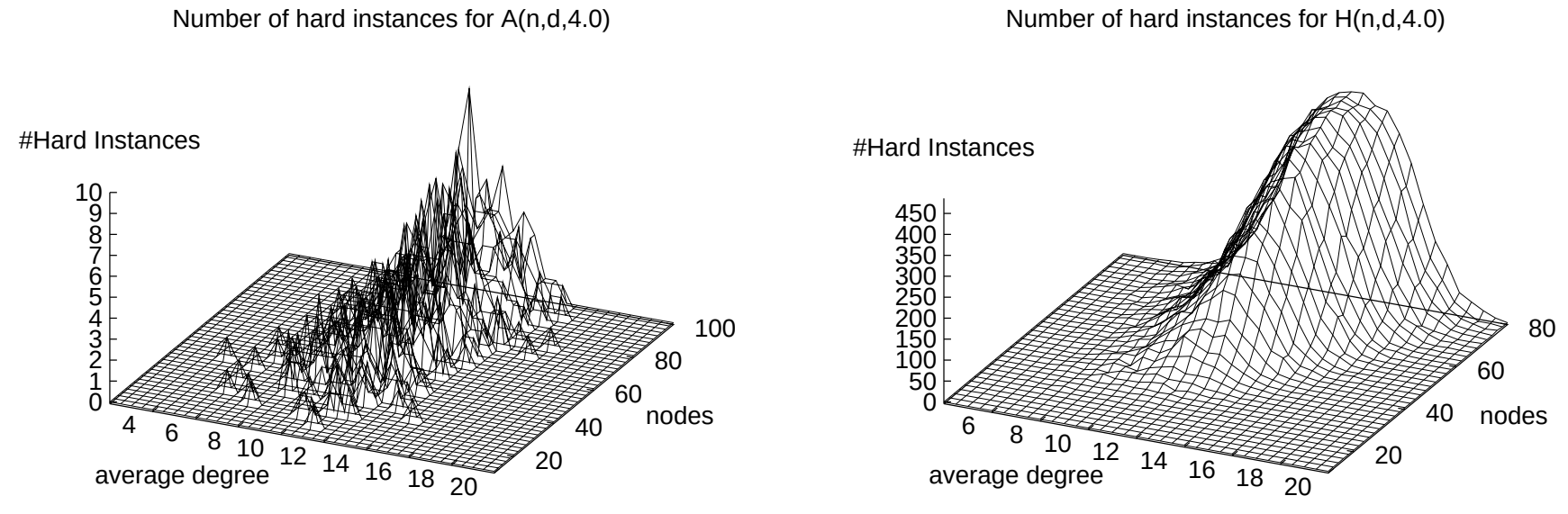

Figure 8: Number of instances using more than 10,000 visited nodes for some heuristic for $A(n, d, 4.0)$ and $H(n, d, 4.0)$

(*static/dynamic* and *local/global* combined with the five split sets $\mathcal{B}, \widehat{\mathcal{B}}, \widehat{\mathcal{H}}_8, \mathcal{C}_8, \mathcal{Q}_8$) on the same randomly generated instances of size $n = 10$ up to $n = 100$. For the instances of $H(n, d, 4.0)$ we restricted ourselves to instances with up to $n = 80$ regions because larger ones appeared to be too difficult.

In first experiments we found that most of the instances were solved very fast with less than 1,000 visited nodes in the search space when using one of the maximal tractable subsets for splitting. However, some instances turned out to be extremely hard, they could not be solved within our limit of 2 million visited nodes, which is about 1.5 hours of CPU time. Therefore, we ran all our programs up to a maximal number of 10,000 visited nodes and stored all instances for which at least one of the different heuristics used more than 10,000 visited nodes for further experiments (see next section). We call those instances the *hard instances*. The distribution of the hard instances is shown in Figure 8. It turned out that for the heuristics using $\mathcal{B}$ as a split set and for the heuristics using *dynamic* and *global* evaluation of the constrainedness many more instances were hard than for the other combinations of heuristics. We, therefore, did not include in Figure 8 the hard instances of the $\mathcal{B}$/*dynamic*/*global* heuristic for $A(n, d, 4.0)$ and the hard instances for the heuristics using $\mathcal{B}$ as a split set and the $\widehat{\mathcal{B}}$/*dynamic*/*global* heuristic for $H(n, d, 4.0)$.

As Figure 8 shows, almost all of the hard instances are in the phase-transition region. For $A(n, d, 4.0)$ only a few of the 500 instances per data point are hard while for $H(n, d, 4.0)$ almost all instances in the phase-transition are hard. Altogether there are 788 hard instances for $A(n, d, 4.0)$ (out of a total number of 759,000 generated instances) and 75,081 hard instances for $H(n, d, 4.0)$ (out of a total number of 594,000 generated instances). Table 1 shows the number of hard instances for each heuristic except for those which were excluded as mentioned above. The heuristics using $\widehat{\mathcal{H}}_8$ as a split set solve more instances than the heuristics using other split sets. Using $\mathcal{C}_8$ or $\mathcal{Q}_8$ as a split set does not seem to be an improvement over using $\widehat{\mathcal{B}}$. Among the different ways of computing constrainedness, *static* and *global* appears to be the most effective combination when using one of the maximal tractable subsets as a split set. For some split sets, *dynamic* and *local* also seems to be an effective combination while combining *dynamic* and *global* is in all cases the worst choice with respect to the number of solved instances.

| Heuristics | $A(n,d,4.0)$ | $H(n,d,4.0)$ | $H(80,14.0,4.0)$ |
|---|---|---|---|
| $\widehat{\mathcal{H}}_8$/sta/loc | 64 | 21,129 | 331 |
| $\widehat{\mathcal{H}}_8$/sta/glo | 42 | 10,826 | 227 |
| $\widehat{\mathcal{H}}_8$/dyn/loc | 52 | 9,967 | 217 |
| $\widehat{\mathcal{H}}_8$/dyn/glo | 100 | 24,038 | 345 |
| $\mathcal{C}_8$/sta/loc | 81 | 28,830 | 373 |
| $\mathcal{C}_8$/sta/glo | 58 | 15,457 | 277 |
| $\mathcal{C}_8$/dyn/loc | 78 | 32,926 | 412 |
| $\mathcal{C}_8$/dyn/glo | 108 | 41,565 | 428 |
| $\mathcal{Q}_8$/sta/loc | 81 | 24,189 | 346 |
| $\mathcal{Q}_8$/sta/glo | 54 | 13,189 | 239 |
| $\mathcal{Q}_8$/dyn/loc | 74 | 13,727 | 255 |
| $\mathcal{Q}_8$/dyn/glo | 104 | 29,448 | 368 |
| $\widehat{\mathcal{B}}$/sta/loc | 68 | 23,711 | 344 |
| $\widehat{\mathcal{B}}$/sta/glo | 89 | 13,831 | 249 |
| $\widehat{\mathcal{B}}$/dyn/loc | 70 | 29,790 | 379 |
| $\widehat{\mathcal{B}}$/dyn/glo | 162 | – | – |
| $\mathcal{B}$/sta/loc | 163 | – | – |
| $\mathcal{B}$/sta/glo | 222 | – | – |
| $\mathcal{B}$/dyn/loc | 209 | – | – |
| $\mathcal{B}$/dyn/glo | (303) | – | – |
| total | 788 | 75,081 | 486 |

Table 1: Number of hard instances for each heuristic

In Figure 9 we compare the 50% and 99% percentiles of the different heuristics on $A(n,d,4.0)$. We do not give the average run times since we ran all heuristics only up to at most 10,000 visited nodes which reduces the real average run time values. Each data point is the average of the values for $d = 8$ to $d = 10$. We took the average of the different degrees in order to cover the whole phase-transition region which is about $d = 8$ for instances of size $n = 10$ and $d = 10$ for instances of size $n = 100$. For all different combinations of computing constrainedness, the ordering of the run times is the same for the different split sets: $\mathcal{B} \gg \widehat{\mathcal{B}} > \mathcal{C}_8, \widehat{\mathcal{H}}_8, \mathcal{Q}_8$. The run times of using *static/local*, *static/global*, or *dynamic/local* for computing constrainedness are almost the same when combined with the same split set while they are longer for all split sets when using *dynamic/global* (about 3 times longer when using $\widehat{\mathcal{B}}$ as a split set and about 1.5 times longer when using the other split sets). The 99% percentile run times are only about 1.5 times longer than the 50% percentile run times. Thus, even the harder among the "normal" instances can be solved easily, i.e., apart from a few hard instances, most instances can be solved efficiently within the size range we analyzed. The erratic behavior of the median curves results from an aggregation of the effect which can be observed in Figure 5, namely, that some of the median elements in the phase-transition are inconsistent and easily solvable.

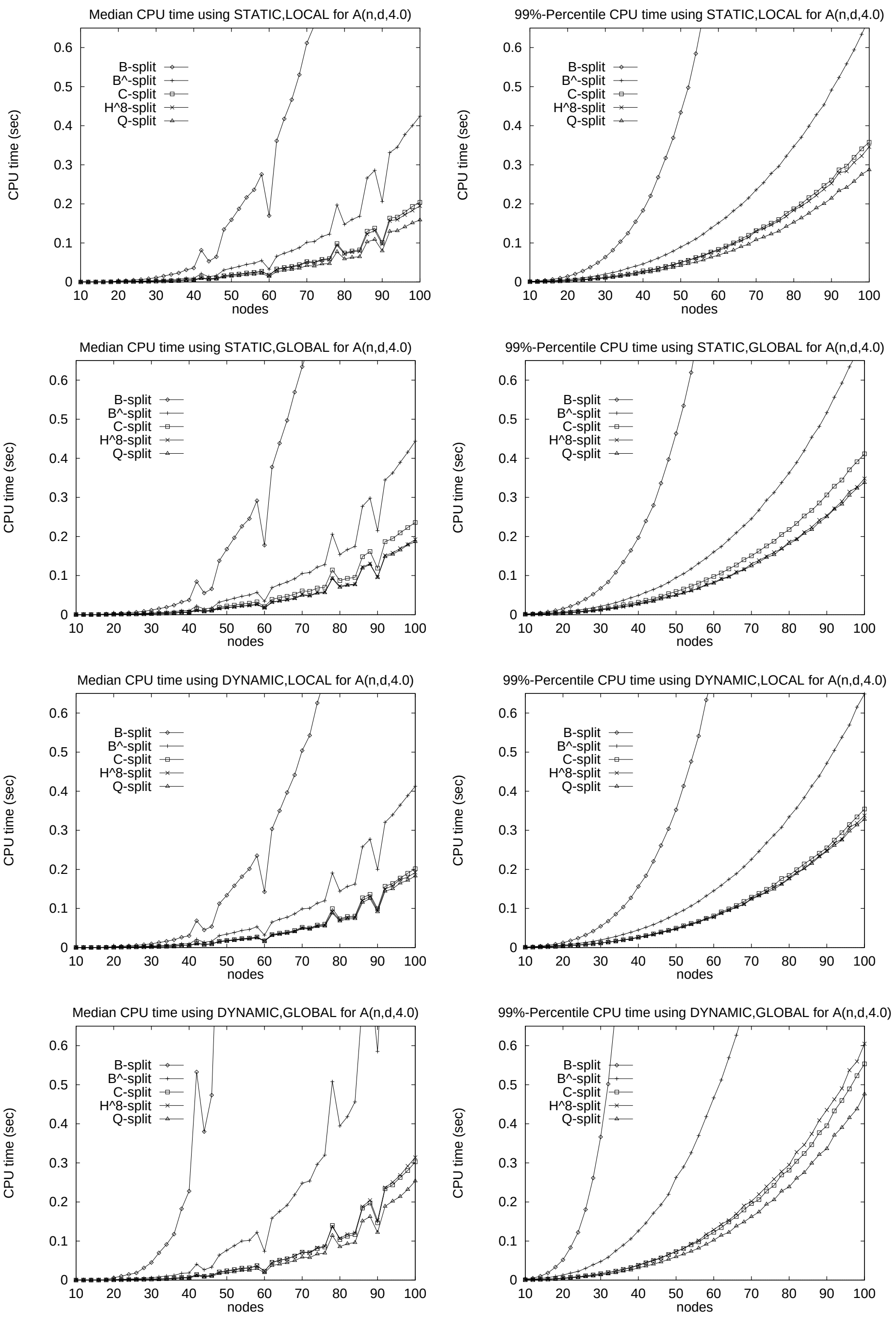

Figure 9: Percentile 50% and 99% CPU time of the different heuristics for solving $A(n, d, 4.0)$ ($d = 8.0$ to $d = 10.0$, 2,500 instances per data point)

For the runtime studies for $H(n, d, 4.0)$ we noticed that there are many hard instances for $n > 40$ (see Figure 8), for $n = 80$ almost all instances in the phase-transition region are hard (see last column of Table 1). Also, as Table 1 shows, the number of hard instances varies a lot for the different heuristics. Therefore, it is not possible to compare the percentile running times of the different heuristics for $n > 40$. For $n = 80$ and $d = 14$ (see last column of Table 1), for instance, the 50% and 99% percentile element of the $\mathcal{C}_8$/*dynamic*/*global* heuristic is element no.36 and element no.72, while it is element no.141 and element no.280 of the $\widehat{\mathcal{H}}_8$/*dynamic*/*local* heuristic (out of the 500 sorted elements), respectively.

For this reason we show the results only up to a size of $n = 40$ (see Figure 10). Again, we took the average of the different degrees from $d = 10$ to $d = 15$ in order to cover the whole phase-transition region. The order of the run times is the same for different combinations of computing constrainedness: $\mathcal{B} \gg \widehat{\mathcal{B}}, \mathcal{C}_8 \gg \mathcal{Q}_8, \widehat{\mathcal{H}}_8$, while $\widehat{\mathcal{H}}_8$ is in most cases the fastest. As for the $A(n, d, 4.0)$ instances, the run times for *dynamic*/*global* were much longer than the other combinations. The 99% percentile run times of the *static*/*global* combination and for $\widehat{\mathcal{H}}_8$ and $\mathcal{Q}_8$ of the *dynamic*/*local* combination are faster than those of the other combinations. Although the median CPU times are about the same as for $A(n, d, 4.0)$ for $n < 40$, the percentile 99% CPU times are much longer. As it was already shown in Figure 7 and 8, this is further evidence that there are very hard instances in the phase-transition region of $H(n, d, 4.0)$.

## 7. Orthogonal Combination of the Heuristics

In the previous section we studied the quality of different heuristics for solving randomly generated RSAT instances. We found that several instances which are mainly located in the phase-transition region could not be solved by some heuristics within our limit of 10,000 visited nodes in the search space. Since the different heuristics have a different search space (depending on the split set) and use a different path through the search space (determined by the different possibilities of computing constrainedness), it is possible that instances are hard for some heuristics but easily solvable for other heuristics. Nebel (1997) observed that running different heuristics in parallel can solve more instances of a particular hard set of temporal reasoning instances proposed by van Beek and Manchak (1996) than any single heuristic alone can solve, when using altogether the same number of visited nodes as for each heuristic alone. An open question of Nebel's investigation (Nebel, 1997) was whether this is also the case for the hard instances in the phase-transition region.

In this section we evaluate the power of "orthogonally combining" the different heuristics for solving RSAT instances, i.e., running the different heuristics for each instance in parallel until one of the heuristics solves the instance. There are different ways for simulating this parallel processing on a single processor machine. One is to use time slicing between the different heuristics, another is to run the heuristics in a fixed or random order until a certain number of nodes in the search space is visited and if unsuccessful try the next heuristic (cf. Huberman, Lukose, & Hogg, 1997). Which possibility is chosen and with which parameters (e.g., the order in which the heuristics are run and the number of visited nodes which is spent for each heuristic) determines the efficiency of the single processor simulation of the orthogonal combination. In order to find the best parameters, we ran all heuristics using at most 10,000 visited nodes for each heuristic on the set of hard instances identified in the previous section (those instances for which at least one heuristic required more than 10,000 visited nodes) and compared their behavior. Since we ran all heuristics on all instances already for the experiments of the previous section, we only had to evaluate their outcomes. This led to a very surprising result for the $A(n,d,4.0)$ instances, namely, all of the 788 hard instances except for a single one were solved by at least one of the heuristics using less than 10,000 visited nodes. In Table 2 we list the percentage of hard instances that could be solved by the different heuristics and the percentage of first response by each of them when running the heuristics in parallel (i.e., which heuristic required the smallest number of visited nodes for solving the instance). It turns out that the heuristics using $\widehat{\mathcal{H}}_8$ as a split set did not only solve more instances than the other heuristics, they were also more often the fastest in finding a solution. Although the heuristics using the other two maximal tractable subsets $\mathcal{Q}_8$ and $\mathcal{C}_8$ as a split set did not solve significantly more instances than the heuristics using $\widehat{\mathcal{B}}$, they were much faster in finding a solution. Despite solving the least number of instances, the heuristics using $\mathcal{B}$ as a split set were in some cases the fastest in producing a solution.

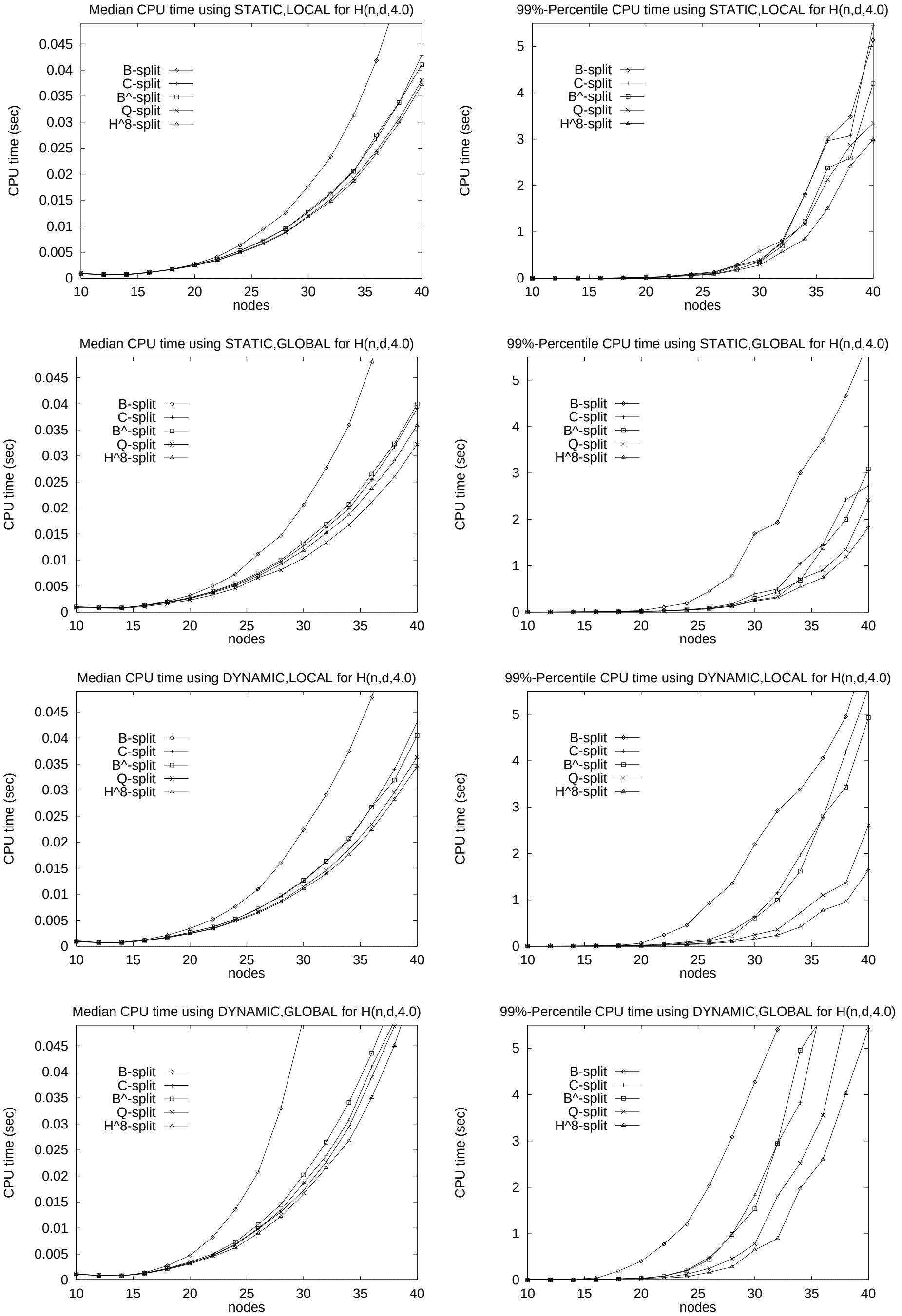

Figure 10: Percentile 50% and 99% CPU time of the different heuristics for solving $H(n, d, 4.0)$ ($d = 10.0$ to $d = 15.0$, 5,500 instances per data point)

| Heuristics | $A(n,d,4.0)$ Solved Instances | $A(n,d,4.0)$ 1. Response | $H(n,d,4.0)$ Solved Instances | $H(n,d,4.0)$ 1. Response |
|---|---|---|---|---|
| $\widehat{\mathcal{H}}_8$/sta/loc | 91.88% | 19.80% | 71.86% | 6.92% |
| $\widehat{\mathcal{H}}_8$/sta/glo | 94.67% | 12.56% | 85.58% | 14.26% |
| $\widehat{\mathcal{H}}_8$/dyn/loc | 93.40% | 24.37% | 86.73% | 22.28% |
| $\widehat{\mathcal{H}}_8$/dyn/glo | 87.31% | 13.58% | 67.98% | 15.00% |
| $\mathcal{C}_8$/sta/loc | 89.72% | 6.35% | 61.60% | 1.47% |
| $\mathcal{C}_8$/sta/glo | 92.64% | 5.20% | 79.41% | 5.04% |
| $\mathcal{C}_8$/dyn/loc | 90.10% | 5.96% | 56.15% | 2.26% |
| $\mathcal{C}_8$/dyn/glo | 86.63% | 6.60% | 44.64% | 2.40% |
| $\mathcal{Q}_8$/sta/loc | 89.72% | 9.77% | 67.78% | 1.63% |
| $\mathcal{Q}_8$/sta/glo | 93.15% | 12.06% | 82.43% | 3.61% |
| $\mathcal{Q}_8$/dyn/loc | 90.61% | 10.15% | 81.72% | 1.83% |
| $\mathcal{Q}_8$/dyn/glo | 86.80% | 12.82% | 60.78% | 4.61% |
| $\widehat{\mathcal{B}}$/sta/loc | 91.37% | 1.40% | 68.42% | 1.84% |
| $\widehat{\mathcal{B}}$/sta/glo | 88.71% | 1.27% | 81.58% | 5.22% |
| $\widehat{\mathcal{B}}$/dyn/loc | 91.12% | 0.89% | 60.32% | 2.56% |
| $\widehat{\mathcal{B}}$/dyn/glo | 79.44% | 0.89% | – | 1.83% |
| $\mathcal{B}$/sta/loc | 79.31% | 0.51% | – | 1.67% |
| $\mathcal{B}$/sta/glo | 71.83% | 0.25% | – | 1.13% |
| $\mathcal{B}$/dyn/loc | 73.48% | 0.51% | – | 0.42% |
| $\mathcal{B}$/dyn/glo | – | 0.13% | – | 0.49% |
| combined | 99.87% | | 96.48% | |

Table 2: Percentage of solved hard instances for each heuristic and percentage of first response when orthogonally running all heuristics. Note that sometimes different heuristics are equally fast. Therefore the sum is more than 100%.

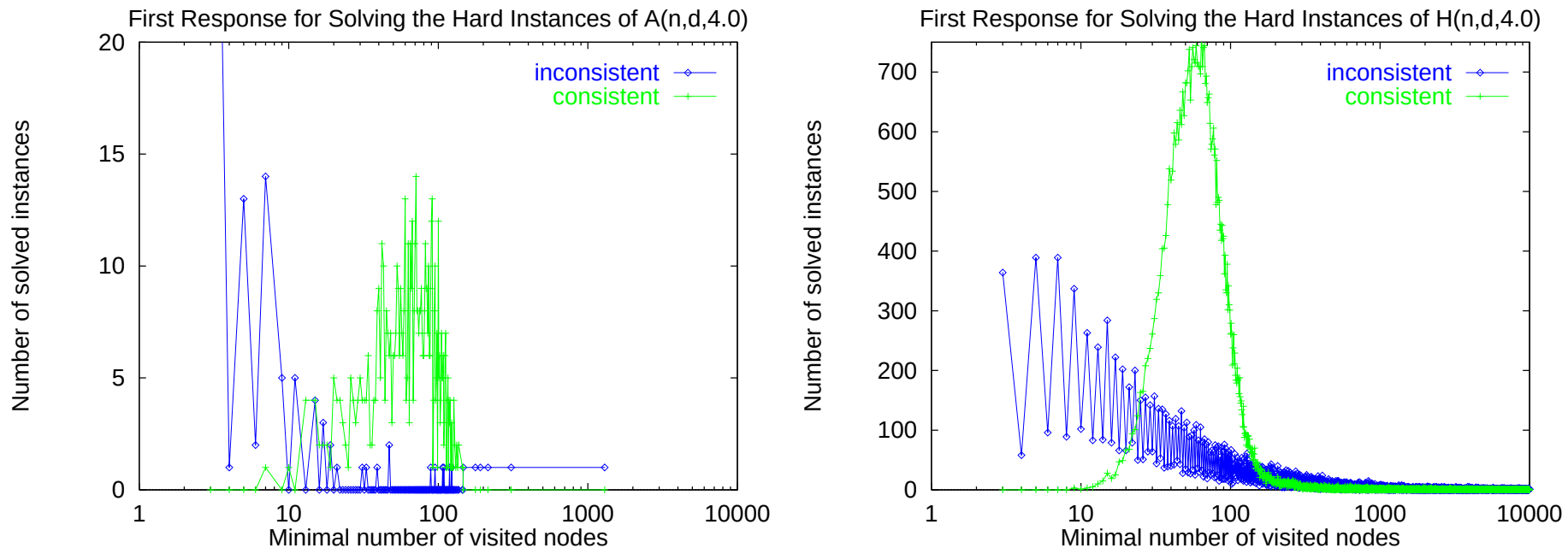

Figure 11: Fastest solution of the hard instances when running all heuristics in parallel

When comparing the minimal number of visited nodes of all the heuristics for all the hard instances, we found that only five of them (which were all inconsistent) required more than 150 visited nodes. This is particularly remarkable as all these instances are from the phase-transition region of an NP-hard problem, i.e., instances which are usually considered to be the most difficult ones. Further note that about 15% (120) of the 788 (path-consistent) instances were inconsistent, which is much higher than usual (cf. Figure 7). Interestingly, most of those inconsistent instances were solved faster than the consistent instances. At this point, it should be noted that combining heuristics orthogonally is very similar to randomized search techniques with restarts (Selman, Levesque, & Mitchell, 1992). However, in contrast to randomized search, our method can also determine whether an instance is inconsistent. In Figure 11 we chart the number of hard instances solved with the smallest number of visited nodes with respect to their solubility. Due to the low number of hard instances of $A(n, d, 4.0)$, the figure on the left looks a bit ugly but one can at least approximate the behavior of the curves when comparing it with the second figure on the right which is the same curve for $H(n, d, 4.0)$ (see below). The oscillating behavior of the inconsistent instances (more instances are solved with an odd than with an even number of visited nodes) might be due to the sizes of the instances—we generated instances with an even number of nodes only. The most difficult instance ($n = 56, d = 10$) was solved as inconsistent with the $\widehat{\mathcal{B}}$/*static*/*global* heuristic using about 91,000 visited nodes while all heuristics using one of the maximal tractable subsets as a split set failed to solve it even when each was allowed to visit 20,000,000 nodes in the search space.

We did the same examination for the set of 75,081 hard instances of $H(n, d, 4.0)$. 2,640 of these instances could not be solved by any of the 20 different heuristics using 10,000 visited nodes each. Their distribution is shown in Figure 12(a). Similar to the hard instances of $A(n, d, 4.0)$, the heuristics using $\widehat{\mathcal{H}}_8$ as a split set were the most successful ones for solving the hard instances of $H(n, d, 4.0)$, as shown in Table 2. They solved more of the hard instances than any other heuristics and produced the fastest response of more than 50% of the hard instances. There is no significant difference between using $\mathcal{C}_8$, $\mathcal{Q}_8$, or $\widehat{\mathcal{B}}$ as a split set, neither in the number of solved instances nor in the percentage of first response. Like in the previous case, computing constrainedness using the *static*/*global* or the *dynamic*/*local* heuristics resulted in more successful paths through the search space by which more instances were solved within 10,000 visited nodes than by the other combinations. On average they produced faster solutions than the other combinations.

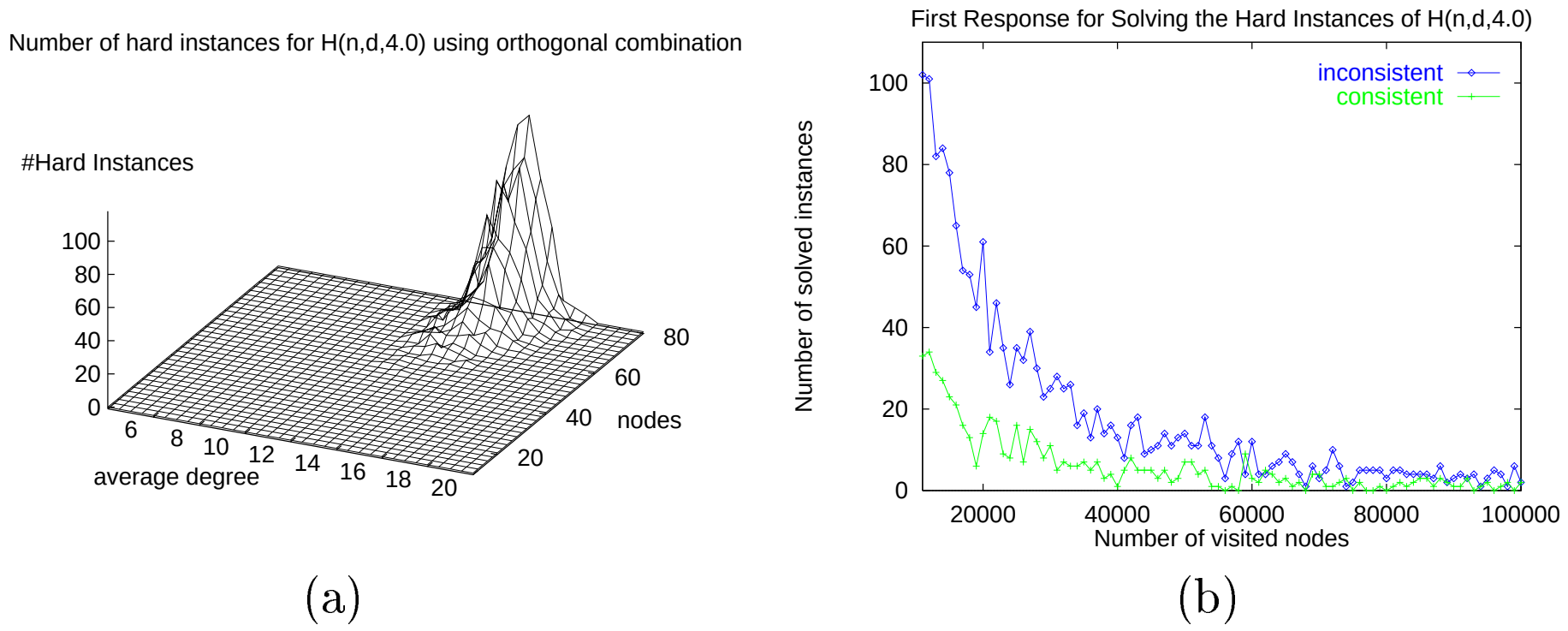

Figure 12: Hard instances using orthogonal combination of all heuristic for $H(n, d, 4.0)$, (a) shows their distribution, (b) shows their fastest solution when using up to 100,000 visited nodes per heuristic

The same observations as for $A(n, d, 4.0)$ can be made when charting the fastest solutions of the hard instances of $H(n, d, 4.0)$ (see Figure 11). About 29% (21,307) of the solved instances are inconsistent. Most of them were, again, solved faster than the consistent instances. More than 75% of the hard instances can be solved with at most 150 visited nodes. 90% can be solved with at most 1,300 visited nodes. Since the $\widehat{\mathcal{H}}_8$/*dynamic*/*local* heuristic alone solves more than 86% of the instances, it seems difficult to combine different heuristics in a way that more hard instances can be solved while using not more than 10,000 visited nodes altogether. However, when orthogonally combining the two best performing heuristics ($\widehat{\mathcal{H}}_8$/*dynamic*/*local* and $\widehat{\mathcal{H}}_8$/*static*/*global*) allowing each of them a maximal number of 5,000 visitable nodes, we can solve 92% (69,056) of the hard instances.

We tried to solve the 2,640 hard instances of $H(n, d, 4.0)$ which are not solvable using orthogonal combination of heuristics with at most 10,000 visited nodes by using a maximal number of 100,000 visited nodes. 471 of these instances are still not solvable, more than 75% of the solved instances are inconsistent. The fastest response for the solved instances is charted in Figure 12(b). The most successful heuristics in giving the fastest response are $\widehat{\mathcal{H}}_8$/*dynamic*/*local* (42.5%) and $\widehat{\mathcal{H}}_8$/*static*/*global* (26.6%). The three heuristics using *static*/*global* computation of constrainedness combined with using $\mathcal{Q}_8, \mathcal{C}_8$, and $\widehat{\mathcal{B}}$ as a split set gave the fastest response for 15.9% of the solved instances where the $\widehat{\mathcal{B}}$ strategy was by far the best among the three (9.4%).

## 8. Combining Heuristics for Solving Large Instances

In the previous section we found that combining different heuristics orthogonally can solve more instances using the same amount of visited nodes than any heuristic alone can solve. In this section we use these results in order to identify the size of randomly generated instances up to which almost all of them, especially those in the phase-transition region, can still be solved in acceptable time. Since many instances of $H(n, d, 4.0)$ are already too difficult for a size of $n = 80$ (see Figure 12), we restrict our analysis to the instances of $A(n, d, 4.0)$ and study randomly generated instances with a size of more than $n = 100$ nodes.

For instances of a large size allowing a maximal number of 10,000 visited nodes in the search space is too much for obtaining an acceptable runtime. 10,000 visited nodes for instances of size $n = 100$ corresponds to a runtime of more than 10 seconds on a Sun Ultra1, for larger instances it gets much slower. Therefore, we have to restrict the maximal number of visited nodes in order to achieve an acceptable runtime. Given a multi-processor machine, the different heuristics can be run orthogonally on different processors using the maximal number of visited nodes each. If the orthogonal combination of the different heuristics is simulated on a single-processor machine, the maximal number of nodes has to be divided by the number of used heuristics to obtain the available number of visitable nodes for each heuristic. Thus, the more different heuristics we use, the less visitable nodes are available for each heuristic. Therefore, in order to achieve the best performance, we have to find the combination of heuristics that solves most instances within a given number of visitable nodes. The chosen heuristics should not only solve many instances alone, they should also complement each other well, i.e., instances which cannot be solved by one heuristic should be solvable by the other heuristic.

We started by finding the optimal combination of heuristics for the set of 788 hard instances of $A(n, d, 4.0)$. From our empirical evaluation given in Section 6 we know how many visited nodes each heuristic needs in order to solve each of the 788 hard instances. Therefore, we computed the number of solved instances for all $2^{20}$ possible combinations of the heuristics using an increasing maximal number of visitable nodes for all heuristics together. Since we only tried to find the combination which solves the most instances, this can be computed quite fast. The results are given in Table 3. They show that a good performance can be obtained with a maximal number of 600 visited nodes. In this case four heuristics were involved, i.e., 150 visitable nodes are spent on each of the four heuristics. Since the same combination of heuristics ($\widehat{\mathcal{H}}_8$/*static*/*global*, $\widehat{\mathcal{H}}_8$/*dynamic*/*local*, $\mathcal{C}_8$/*dynamic*/*local*, $\widehat{\mathcal{B}}$/*static*/*local*) is also the best for up to 1,000 visitable nodes, we choose this combination for our further analysis. We choose the order in which they are processed to be 1. $\widehat{\mathcal{H}}_8$/*dynamic*/*local*, 2. $\widehat{\mathcal{H}}_8$/*static*/*global*, 3. $\mathcal{C}_8$/*dynamic*/*local*, 4. $\widehat{\mathcal{B}}$/*static*/*local* according to their first response behavior given in Table 2. Note that although the two heuristics $\mathcal{C}_8$/*dynamic*/*local* and $\widehat{\mathcal{B}}$/*static*/*local* do not show a particularly good performance when running them alone (see Table 2), they seem to best complement the other two heuristics.

What we have to find next is the maximal number of visitable nodes we spend for the heuristics. For this we ran the best performing heuristic ($\widehat{\mathcal{H}}_8$/*dynamic*/*local*) on instances of the phase-transition region of varying sizes. It turned out that for almost all consistent instances the number of visited nodes required for solving them was slightly less than twice the size of the instances while most inconsistent instances are also not path-consistent and, thus, solvable with only one visited node. Therefore, we ran the four heuristics in the following allowing $2n$ visited nodes each, where $n$ is the size of the instance, i.e., together we allow at most $8n$ visitable nodes. We randomly generated test instances according to the $A(n, d, 4.0)$ model for a size of $n = 110$ regions up to a size of $n = 500$ regions with a step of 10 regions and 100 instances for each size and each average degree ranging from $d = 2.0$ to $d = 18.0$ with a step of 0.5, a total number of 132,000 instances. Since solving large instances using backtracking requires a lot of memory, we solved the instances on a Sun Ultra60 with 1GB of main memory.

| Max Nodes | Solved Instances | Combination of Heuristics |
|---|---|---|
| 100 | 516 | $\widehat{\mathcal{H}}_8$-d-l |
| 200 | 705 | $\widehat{\mathcal{H}}_8$-s-g |
| 300 | 759 | $\widehat{\mathcal{H}}_8$-s-g, $\widehat{\mathcal{H}}_8$-d-l |
| 400 | 769 | $\widehat{\mathcal{H}}_8$-s-g, $\mathcal{C}_8$-d-l |
| 500 | 774 | $\widehat{\mathcal{H}}_8$-s-g, $\widehat{\mathcal{H}}_8$-d-l, $\mathcal{C}_8$-d-l |
| 600 | 778 | $\widehat{\mathcal{H}}_8$-s-g, $\widehat{\mathcal{H}}_8$-d-l, $\mathcal{C}_8$-d-l, $\widehat{\mathcal{B}}$-s-l |
| 700 | 780 | $\widehat{\mathcal{H}}_8$-s-g, $\widehat{\mathcal{H}}_8$-d-l, $\mathcal{C}_8$-d-l, $\widehat{\mathcal{B}}$-s-l |
| 800 | 783 | $\widehat{\mathcal{H}}_8$-s-g, $\widehat{\mathcal{H}}_8$-d-l, $\mathcal{C}_8$-d-l, $\widehat{\mathcal{B}}$-s-l |
| 900 | 784 | $\widehat{\mathcal{H}}_8$-s-g, $\widehat{\mathcal{H}}_8$-d-l, $\mathcal{C}_8$-d-l, $\widehat{\mathcal{B}}$-s-l |
| 1100 | 785 | $\widehat{\mathcal{H}}_8$-s-g, $\widehat{\mathcal{H}}_8$-d-l, $\mathcal{C}_8$-d-l, $\widehat{\mathcal{B}}$-s-l, $\widehat{\mathcal{B}}$-s-g |
| 1300 | 786 | $\widehat{\mathcal{H}}_8$-s-g, $\widehat{\mathcal{H}}_8$-d-l, $\widehat{\mathcal{B}}$-s-l, $\widehat{\mathcal{B}}$-s-g |
| 3900 | 787 | $\widehat{\mathcal{H}}_8$-s-g, $\widehat{\mathcal{H}}_8$-d-l, $\widehat{\mathcal{B}}$-d-l |

Table 3: Best performance of combining different heuristics for solving the 787 solvable hard instances of $A(n, d, 4.0)$ with a fixed maximal number of visited nodes

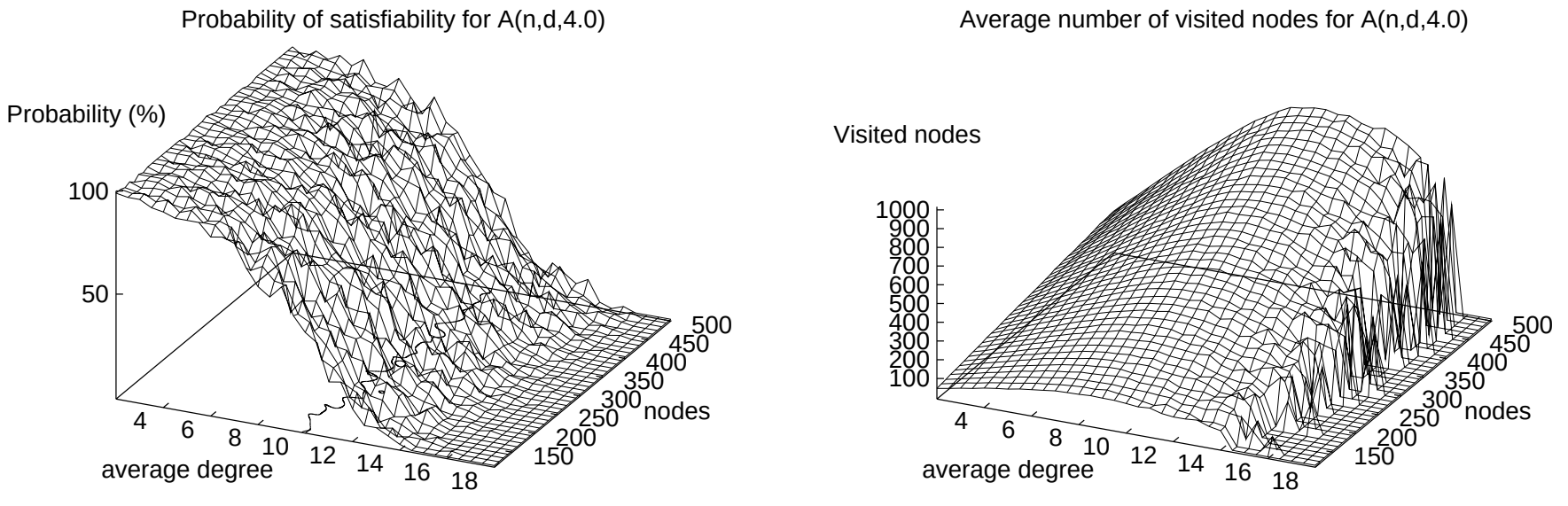

Figure 13: Probability of satisfiability for $A(n, d, 4.0)$ (100 instances per data point) and average number of visited nodes of the path-consistent instances when using orthogonal combination of the four selected heuristics

The generated instances display a phase-transition behavior which continues the one given in Figure 5. The phase-transition ranges from $d = 10.0$ for $n = 110$ to $d = 10.5$ for $n = 500$ (see Figure 13). Apart from 112 instances, all other instances we generated were solvable by orthogonal combination of the four heuristics ($\widehat{\mathcal{H}}_8$/*static*/*global*, $\widehat{\mathcal{H}}_8$/*dynamic*/*local*, $\mathcal{C}_8$/*dynamic*/*local*, $\widehat{\mathcal{B}}$/*static*/*local*) spending less than 2n visited nodes

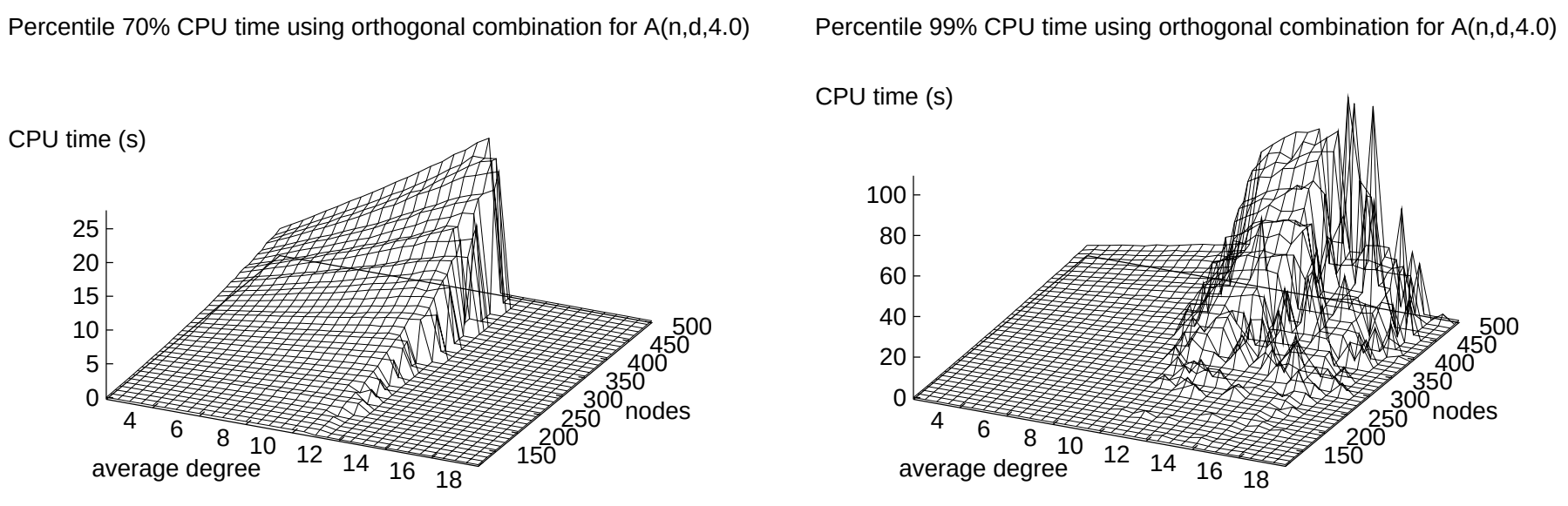

Figure 14: Percentile 70% and 99% CPU time of the orthogonal combination of four different heuristics for solving large randomly generated instances of $A(n, d, 4.0)$

each. In Figure 13 we give the average number of visited nodes of the path-consistent instances. It can be seen that for our test instances the average number of visited nodes is linear in the size of the instances. The percentile 70% CPU time for instances of the phase-transition with a size of $n = 500$ regions is about 20 seconds, the percentile 99% CPU time is about 90 seconds. Up to a size of $n = 400$ regions, the percentile 99% CPU time is less than a minute (see Figure 14).

131,240 of our test instances were already solved by the $\widehat{\mathcal{H}}_8$/*static*/*global* heuristic, for 71 instances the $\widehat{\mathcal{H}}_8$/*dynamic*/*local* heuristic was required and for 577 instances the $\mathcal{C}_8$/*dynamic*/*local* heuristic produced the solution. None of the 112 instances which were not solved by one of those three heuristics were solved by the $\widehat{\mathcal{B}}$/*static*/*local* heuristic. We tried to solve these instances using the other heuristics, again using a maximal number of $2n$ visited nodes each. The best performing among those heuristics was the $\mathcal{C}_8$/*dynamic*/*global* heuristic which solved 87 of the 112 instances followed by the $\mathcal{C}_8$/*static*/*global* heuristic (83) and the $\mathcal{Q}_8$/*dynamic*/*global* heuristic (63). 7 instances were not solved by any heuristic within a maximal number of $2n$ visited nodes.

## 9. Discussion

We empirically studied the behavior of solving randomly generated RSAT instances using different backtracking heuristics some of which make use of the maximal tractable subsets identified in previous work. We generated instances according to two different models of which the "general model" $A$ allows all 256 RCC-8 relations to be used while the "hard model" $H$ allows only relations which are not contained in any of the maximal tractable subsets. A theoretical analysis of the two models showed that the model $H$ and the model $A$ for a small average degree of the nodes in the constraint graph do not suffer from trivial local inconsistencies as it is the case for similar generation procedures for CSPs with finite domains (Achlioptas et al., 1997). It turned out that randomly generated instances of both models show a phase-transition behavior which depends most strongly on the average degree of the instances. While most instances outside the phase-transition region can be solved efficiently by each of our heuristics, instances in the phase-transition region can be extremely hard. For the instances of the general model, most path-consistent instances are also consistent. Conversely, path-consistency is a bad approximation to consistency for instances of the hard model. These instances are also much harder to solve than instances of the general model.

When comparing the different heuristics, we found that the heuristics using one of the maximal tractable subsets as a split set are not as much faster in deciding consistency of RSAT instances as their theoretical advantage given by the reduced average branching factor and the resulting exponentially smaller size of the search space indicates. This is because using path-consistency as a forward checking method considerably reduces the search space in all cases. Nevertheless, using one of the maximal tractable subsets as a split set, in particular $\widehat{\mathcal{H}}_8$, still leads to a much faster solution and solves more instances in reasonable time than the other heuristics. Although the two maximal tractable subsets $\mathcal{Q}_8$ and $\mathcal{C}_8$ contain more relations than $\widehat{\mathcal{H}}_8$, their average branching factor is lower, i.e., when using $\widehat{\mathcal{H}}_8$ one has to decompose more relations ($256 - 148 = 108$) than when using the other two sets (96 and 98 relations, respectively), but $\widehat{\mathcal{H}}_8$ splits the relations better than the other two sets. Most relations can be decomposed into only two $\widehat{\mathcal{H}}_8$ sub-relations, while many relations must be decomposed into three $\mathcal{C}_8$ sub-relations or into three $\mathcal{Q}_8$ sub-relations. This explains the superior performance of heuristics involving $\widehat{\mathcal{H}}_8$ for decomposition.

Among the instances we generated, we stored those which could not be solved by all heuristics within a maximum number of 10,000 visited nodes in the search space in order to find out how the different heuristics perform on these hard instances. We found that almost all hard instances are located in the phase-transition region and that there are many more hard instances in the hard model than in the general model. We orthogonally combined all heuristics and ran them on all hard instances. This turned out to be very successful. Apart from one instance, all hard instances of the general model could be solved, most of them with a very low number of visited nodes. The hard instances of the hard model were much more difficult: many of them could not be solved by any of the heuristics. Nevertheless, many more instances were solved by orthogonally combining the heuristics than by each heuristic alone. Again, most of them were solved using a low number of visited nodes.

Based on our observations on orthogonally combining different heuristics, we tried to identify the combination of heuristics which is most successful in efficiently solving many instances and used this combination for solving very large instances. It turned out that the best combination involves only heuristics which use maximal tractable subsets for decomposition. With this combination we were able to solve almost all randomly generated instances of the phase-transition region of the general model up to a size of $n = 500$ regions very efficiently. This seems to be impossible when considering the enormous size of the search space, which is on average $10^{39323}$ for instances of size $n = 500$ when using $\widehat{\mathcal{H}}_8$ as a split set.

Our results show that despite its NP-hardness, we were able to solve almost all randomly generated RSAT instances of the general model efficiently. This is neither due to the low number of different RCC-8 relations (instances generated according to the hard model are very hard in the phase-transition region) nor to our generation procedure for random instances which does not lead to trivially flawed instances asymptotically. It is mainly due to the maximal tractable subsets which cover a large fraction of RCC-8 and which lead to extremely low branching factors. Since there are different maximal tractable subsets, they allow choosing between many different backtracking heuristics which further increases efficiency: some instances can be solved easily by one heuristic, other instances by other heuristics. Heuristics involving maximal tractable subclasses showed the best behavior but some instances can be solved faster when other tractable subsets are used. The full classification of tractable subsets gives the possibility of generating hard instances with a high probability. Many randomly generated instances of the phase-transition region are very hard when using only relations which are not contained in any of the tractable subsets and consist of more than $n = 60$ regions. The next step in developing efficient reasoning methods for RCC-8 is to find methods which are also successful in solving most of the hard instances of the hard model.

The results of our empirical evaluation of reasoning with RCC-8 suggest that analyzing the computational properties of a reasoning problem and identifying tractable subclasses of the problem is an excellent way for achieving efficient reasoning mechanisms. In particular maximal tractable subclasses can be used to develop more efficient methods for solving the full problem since their average branching factor is the lowest. Using the refinement method developed in Renz's (1999) paper, tractable subclasses of a set of relations forming a relation algebra can be identified almost automatically. This method makes it very easy to develop efficient algorithms. A further indication of our empirical evaluation is that it can be much more effective (even and especially for hard instances of the phase-transition region) to orthogonally combine different heuristics than to try to get the final epsilon out of a single heuristic. This answers a question raised by Nebel (1997) of whether the orthogonal combination of heuristics is also useful in the phase-transition region. In our experiments this lead to much better results even when simulating the orthogonal combination of different heuristics on a single processor machine and spending altogether the same resources as for any one heuristic alone. In contrast to the method of time slicing between different heuristics, we started a new heuristic only if the previous heuristic failed after a certain number of visited nodes in the search space. The order in which we ran the heuristics depended on their performance and on how well they complemented each other, more successful heuristics were used first. This is similar to using algorithm portfolios as proposed by Huberman et al. (1997). Which heuristics perform better and which combination is the most successful one is a matter of empirical evaluation and depends on the particular problem. Heuristics depending on maximal tractable subclasses, however, should lead to the best performance.

For CSPs with finite domains there are many theoretical results about localizing the phase-transition behavior and about predicting where hard instances are located. In contrast to this, there are basically no such theoretical results for CSPs with infinite domains as used in spatial and temporal reasoning. As our initial theoretical analysis shows, theoretical results on CSPs with finite domains do not necessarily extend to CSPs with infinite domains. It would be very interesting to develop a more general theory for CSPs with infinite domains, possibly similar to Williams and Hogg's "Deep Structure" (Williams & Hogg, 1994) or Gent et al.'s "Kappa" theory (Gent, MacIntyre, Prosser, & Walsh, 1996).

## Acknowledgments

We would like to thank Ronny Fehling for his assistance in developing the programs, Malte Helmert for proof reading the paper, and the three anonymous reviewers for their very helpful comments.
This research has been supported by DFG as part of the project FAST-QUAL-SPACE, which is part of the DFG special research effort on "Spatial Cognition". The first author has been partially supported by a Marie Curie Fellowship of the European Community programme "Improving Human Potential" under contract number HPMF-CT-2000-00667. A preliminary version of this paper appeared in the *Proceedings of the 13th European Conference on Artificial Intelligence* (Renz & Nebel, 1998).